\documentclass{article}
\usepackage{spconf,amsmath,graphicx}
\usepackage[hidelinks]{hyperref}       
\usepackage{url}            
\usepackage{booktabs}       
\usepackage{amsfonts}       
\usepackage{xspace}

\usepackage{mathtools}
\usepackage{paralist}
\usepackage{algorithmicx}
\usepackage{algpseudocode}  
\usepackage{algorithm}
\usepackage{diagbox}
\usepackage{multirow}
\usepackage{subfigure}
\usepackage{color}
\usepackage{caption}

\title{Universal Soldier: Using Universal Adversarial Perturbations for Detecting Backdoor Attacks}

\name{Xiaoyun Xu$^{\star}$ \qquad Oguzhan Ersoy$^{\star}$ \qquad Hamidreza Tajalli$^{\star}$ \qquad Stjepan Picek$^{\star}$}

\address{$^{\star}$ Digital Security Group, Radboud University, The Netherlands\\
\{xiaoyun.xu, oguzhan.ersoy, hamidreza.tajalli, stjepan.picek\}@ru.nl}

\begin{document}
%
\maketitle

\DeclarePairedDelimiterX{\set}[1]{\{}{\}}{\setargs{#1}}
\NewDocumentCommand{\setargs}{>{\SplitArgument{1}{;}}m}
{\setargsaux#1}
\NewDocumentCommand{\setargsaux}{mm}
{\IfNoValueTF{#2}{#1} {#1\,\delimsize|\,\mathopen{}#2}}

\newcommand{\norm}[1]{$\lVert$ #1 $\rVert$}

\newcommand{\ourmethod}{\texttt{USB}\xspace}

\begin{abstract}
This paper proposes a backdoor detection method by utilizing a particular type of adversarial attack, universal adversarial perturbation (UAP), and its similarities with a backdoor trigger.
We observe an intuitive phenomenon: UAPs generated from backdoored models need fewer perturbations to mislead the model than UAPs from clean models.
UAPs of backdoored models tend to exploit the shortcut from all classes to the target class, built by the backdoor trigger.
We propose a novel method called Universal Soldier for Backdoor detection (\ourmethod) and reverse engineering potential backdoor triggers via UAPs.
Experiments on 240 models trained on several datasets show that \ourmethod effectively detects the injected backdoor and provides comparable or better results than state-of-the-art methods.
\end{abstract}
\section{Introduction}
\label{sec:introduction}
\begin{figure}[t]
\centering
\includegraphics[width = 0.8\linewidth]{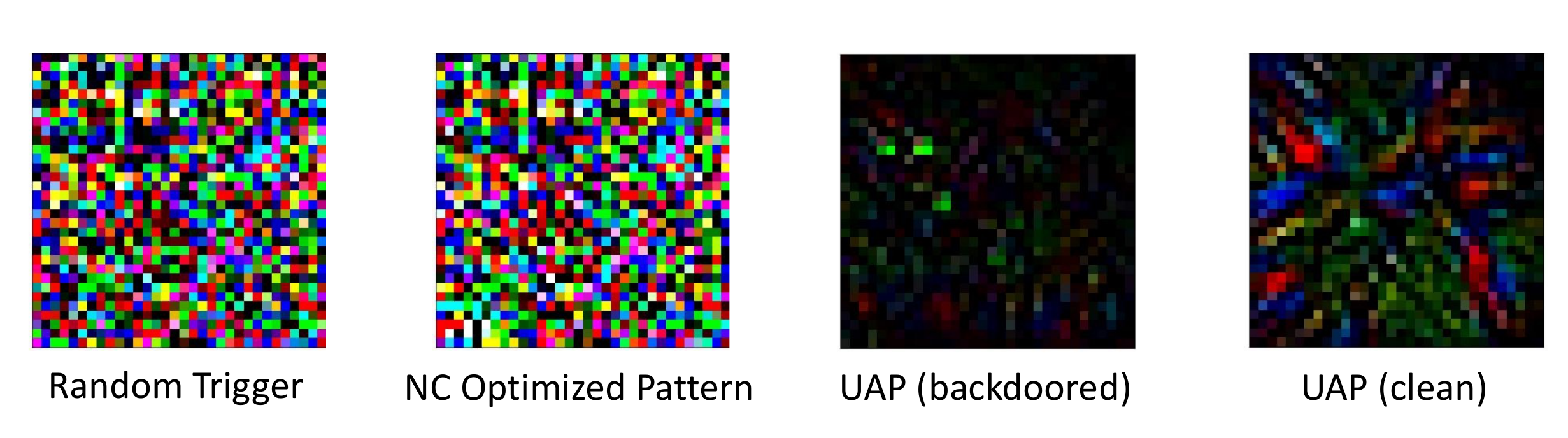}
\caption{The random point is barely updated by NC.}
\label{fig:startpoints}
\end{figure}

There have been several proposals to detect backdoor attacks~\cite{8835365,10.1145/3319535.3363216,9338311} by analyzing the pre-trained model. 
In particular, reverse engineering, such as Neural Cleanse (NC)~\cite{8835365} and TABOR~\cite{9338311},
aims to reconstruct the trigger through for every class.
The backdoored class behaves differently compared to clean classes.
However, previous methods may capture the unique feature of the target class (for example, wings of the airplane class) instead of the trigger~\cite{10.1145/3319535.3363216}.
Both class features and triggers can lead a backdoored model to the target class.
Reverse engineering decides whether there is a backdoor according to the size (for example, $L_1 norm$) of reconstructed triggers of every class.
If the difference between the unique class feature and the trigger is not particularly large concerning size, reverse engineering may not generate the trigger, see Fig.~\ref{fig:classfeaturetriggers} as an example.
Furthermore, these methods work well against patch based triggers, such as BadNet~\cite{8685687}, but may fail under non-patch-wised attacks (see Tab.~\ref{tab:cifarvgg16otherattacks}), such as Input-aware dynamic
backdoor attack (IAD)~\cite{NEURIPS2020_234e6913}.
The reason is that reverse engineering usually starts from a random point. 
The content in random points can be very different from triggers designed by more advanced attacks, and NC-style methods only optimize the mask without directly updating trigger patterns, see Fig.~\ref{fig:startpoints}. 
Therefore, it is difficult for NC-style methods to generate specifically designed triggers.
In addition, NC-style methods also need lots of data to perform the optimization with a larger number of iterations.

This paper presents a novel detection mechanism that does not suffer from the aforementioned concerns.
More specifically, we investigate an inference-time defense that only requires a small amount of clean data.
To avoid using the class unique feature as a trigger, we utilize the similarities between backdoor attacks and adversarial attacks, especially universal adversarial perturbations (UAP)~\cite{Moosavi-Dezfooli_2017_CVPR}.
The UAP effectively fools the victim model on any inputs because it captures the correlations among different regions of the decision boundary~\cite{Moosavi-Dezfooli_2017_CVPR}.
With similar reasoning, we conjecture that UAP can also capture the feature of backdoor neurons, resulting in slighter perturbations, see Fig.~\ref{fig:startpoints}.
\ourmethod requires less iteration and data as we directly use UAP to capture the potential backdoor.
As UAP can be generalized across different networks, we only need to generate UAP once for similar models, greatly reducing the time requirement.

\ourmethod is evaluated on 240 models (150 on CIFAR-10~\cite{krizhevsky2009learning} by ResNet-18~\cite{He_2016_CVPR}, 45 on ImageNet~\cite{5206848} by Efficientnet-B0~\cite{pmlr-v97-tan19a}, 45 on CIFAR-10~\cite{krizhevsky2009learning} by VGG-16~\cite{DBLP:journals/corr/SimonyanZ14a} with stronger attacks), and \ourmethod achieves better backdoor detection performance than previous methods on these datasets and architectures. 
Our main contributions are summarized as follows:
\begin{compactitem}
    \item We propose a novel detection method \ourmethod that utilizes the similarities between backdoor and UAPs.
    \item Our method \ourmethod can detect stronger backdoor triggers for both patch-wised (BadNet and Latent) and non-patch-wised (Input Aware Dynamic) backdoors. 
    \item We conduct experiments on 240 models to evaluate the effectiveness and limitations of our method. 
\end{compactitem}

\section{Preliminaries}
\label{sec:related}
\noindent\textbf{Inference-time Defenses.}
Inference-time defenses refer to defenses with access to the pre-trained model and a certain amount of clean data, including detection by reverse engineering of the backdoor trigger~\cite{8835365,10.1145/3319535.3363216,Dong_2021_ICCV}, pruning~\cite{NEURIPS2021_8cbe9ce2}, and machine unlearning to remove the backdoor~\cite{8835365}.
The pruning and machine unlearning aim to remove the backdoor by directly modifying the victim model, while the reverse engineering conducts detection and reconstructs the backdoor trigger.
Reverse engineering methods, such as NC~\cite{8835365} and TABOR~\cite{9338311}, use the behavioral characteristics of the backdoor itself.
The backdoor builds a shortcut from within regions of the space belonging to each label into the region belonging to the target.
For backdoored models, transforming input features of any class into features of the target class requires less perturbation than transforming into clean classes.

\renewcommand{\algorithmicrequire}{\textbf{Input:}}
\renewcommand{\algorithmicensure}{\textbf{Output:}}
\begin{algorithm}[t]
\footnotesize
	\caption{Computation of targeted UAP.}
	\begin{algorithmic}[1]
		\Require{Data points $X$, target class $t$, victim model $f$, desired $l_p$ norm of the perturbation $\delta$, desired error rate $\theta$}
		\Ensure{Targeted UAP $v$}
		\State Initialize $v \leftarrow 0$
		\While{Err($X+v$) $\leq$ $e$}
		\For{$x_i$ in $X$}
		\If{ $f(x_i + v) \neq t $ }
		\State Minimal perturbation that send $x_i + v$ to class $t$:
		\State $\triangle v_i \leftarrow arg \mathop{min}\limits_{r} \lVert r \rVert_2 \ s.t. f(x_i+v+r) = t $
		\State $v \leftarrow \triangle v_i $ \Comment{Update the perturbation under limitation}
		\EndIf
		\EndFor
		\EndWhile
	\end{algorithmic} 
	\label{alg:tuap}
\end{algorithm}

\section{Proposed Method}
\label{sec:method}

\subsection{Threat Model.}
We consider defense against a backdoor attack under the  Machine Learning as a Service (MLaaS) scenario.
In MLaaS, users with insufficient computation and training resources resort to remote computation to train high-performance models.
The adversary (attacker) controls (or has access to) the MLaaS platform and aims to inject backdoors into models for malicious purposes.
The defender aims to detect these backdoors.
In this paper, we consider all-to-one backdoor attacks.


\begin{algorithm}[t]
\footnotesize
	\caption{Updating of targeted UAP.} 
	\begin{algorithmic}[1]
		\Require{Data points $X$, target class $t$, victim model $f$, UAP $v$, Maximum iteration number $m$, learning rate $lr$}
		\Ensure{Updated UAP $v' = trigger \times mask$}
		\State Initialize trigger and mask by $v: trigger \times mask = v$
		\For{$i = 0$ to $m$}
            \State $x \subseteq X$ \Comment{Take a batch of data, $x$, from $X$ in order}
		\State $x' = x \times (1 - mask) + trigger \times mask$ \Comment{Apply trigger and mask to get perturbed input}
		\State $output = f(x')$
		\State $\mathcal{L}=\mathcal{L}(output, t) - SSIM(x, x') + norm_{L1}(mask)$ 
		\State Backward loss $\mathcal{L}$ to update $mask$ and $trigger$
		\State $mask: mask \leftarrow mask - lr \times \nabla mask$
		\State $trigger: trigger \leftarrow trigger - lr \times \nabla trigger$
            \State $v' = trigger \times mask $
		\EndFor
	\end{algorithmic} 
	\label{alg:updateuap}
\end{algorithm}

\subsection{Targeted UAP}
To work in all-to-one situation, we modify the algorithm from~\cite{Moosavi-Dezfooli_2017_CVPR} to generate targeted UAP.
We formalize the algorithm, i.e., Alg.~\ref{alg:tuap}.
Let us assume a pre-trained deep learning model $f$ and $K$ entries of training data $D=\set{(x_i,y_i)}_{0}^{K-1}$ where $x_i \in \mathbb{R}^{d_X}$ and $y_i \in \set{0,1}^{N}$.
N is the number of classes, and $d_X$ is the input dimension.
The objective of the targeted UAP algorithm is to find a perturbation vector $v$ that misleads the model $f$ on most of the data points in $D$ to a target class $t$.
To work in a more realistic situation, we use a very small number of data points $X$ sampled from the same distribution as $D$ for the algorithm.
Empirically, a size smaller than 1\% of $D$ can be enough for $X$.

The details are described in Alg.~\ref{alg:tuap}. The algorithm iteratively goes through every data point in $X$ to update UAP from scratch.
At each iteration, the algorithm searches for the minimal perturbation that sends $x_i + v$ to the target class.
Then, the error rate of inputting $X+v$ to $f$ should be larger than the desired threshold $\theta$.
This is feasible by solving the following optimization problem:
\begin{equation*}
	\triangle v_i \leftarrow arg \mathop{min}\limits_{r} \lVert r \rVert_2 \ s.t. f(x_i+v+r) = t.
\end{equation*}
Following the algorithm in~\cite{Moosavi-Dezfooli_2017_CVPR}, this search optimization is implemented by deepfool~\cite{Moosavi-Dezfooli_2016_CVPR}.

\subsection{UAP Optimization}
Targeted UAP might be enough to contain the feature related to the backdoor trigger.
Then we further analyze the potential trigger by  an optimization phase to update the targeted UAP.
The optimization objective is formalized as a loss function:
\begin{equation}
	\mathcal{L} = \mathcal{L}_{ce}(output, t) - SSIM(x, x') + norm_{L1}(mask),
 \label{eq:optimizationloss}
\end{equation}
where $\mathcal{L}_{ce}$ refers to the cross-entropy loss.
The structural similarity index measure (SSIM) is a measure of the similarity between images~\cite{1284395}.

The details are provided in Alg.~\ref{alg:updateuap}.
The optimization achieves two goals: (1) make the targeted UAP focus on more important pixels, and (2) ensure that the UAP can mislead the victim model.
As mentioned before, misleading a backdoored model to the target class needs smaller perturbation compared to untarget classes.
Therefore, if $f$ is backdoored on class $t_b$, the size of $v_{t_b}'$ will be smaller than other UAPs in $\set{v_i'}_0^{N-1}$.
For example, for a ResNet-18 model with a BadNet backdoor on class 0, the $L_1$ norm $v_0'$ generated by \ourmethod is 4.49, and
the average $L_1$ norm of the other classes is 53.76.

\newcommand{\tabincell}[2]{\begin{tabular}{@{}#1@{}}#2\end{tabular}}

\begin{table*}[ht]
\footnotesize
\centering
\setlength\tabcolsep{2pt}
\begin{tabular}{cccccccccccc}
\toprule
\multirow{2}{*}{Model} & \multirow{2}{*}{Accuracy} & \multirow{2}{*}{ASR} & \multirow{2}{*}{Method} & \multicolumn{1}{c}{Reversed Trigger} & \multicolumn{2}{c}{Model Detection} & \multicolumn{3}{c}{Target Class Detection} \\
~& ~& ~& ~& $L_1$ norm & Clean & Backdoored & Correct  & Correct Set & Wrong\\
\midrule
\multirow{3}{*}{Clean} & \multirow{3}{*}{85.38} & \multirow{3}{*}{N/A} & NC & 51.59 & 50  & 0  & N/A & N/A & N/A \\
~& ~& ~& TABOR& 55.09 & 50 & 0  & N/A & N/A & N/A \\
~& ~& ~& \ourmethod& 48.99 & 50  & 0  & N/A & N/A & N/A \\
\midrule
\multirow{3}{*}{\tabincell{c}{Backdoored \\ (2 $\times$ 2 trigger)}} & \multirow{3}{*}{83.43} & \multirow{3}{*}{95.04} & NC & 8.72 & 5  & 45  & 44 & 1 & 0 \\
~& ~& ~& TABOR& 9.26 & 5  & 45  & 44 & 1 & 0 \\
~& ~& ~& \ourmethod& 9.83 & 1  & 49  & 45  & 4  & 0  \\
\midrule
\multirow{3}{*}{\tabincell{c}{Backdoored \\ (3 $\times$ 3 trigger)}} & \multirow{3}{*}{83.59} & \multirow{3}{*}{97.57} & NC & 8.89 & 2  & 48  & 48 & 0 & 0\\
~& ~& ~& TABOR& 10.06 & 3 & 47  & 47 & 0 & 0\\
~& ~& ~& \ourmethod& 12.02 & 1  & 49  & 49  & 0  & 0  \\
\bottomrule
\end{tabular}
\caption{Detection evaluation on CIFAR-10 where each case consists of  50 trained models.}
\label{tab:cifar10detection}
\end{table*}

\section{Evaluation}
\label{sec:experiments}

In this section, we provide the experimental results for \ourmethod as well as its comparison with NC~\cite{8835365} and TABOR~\cite{9338311}, which are the typical and state-of-the-art methods.
Experiments are conducted with TrojanZoo~\cite{pang:2022:eurosp}, as they provide a widely-used implementation of various attacks and defense technologies.
We use different random seeds for every trained model.

\begin{figure}[t]
\footnotesize
    \centering
      \subfigure{
      \rotatebox{90}{~~\scriptsize{CIFAR-10}}
		\begin{minipage}[b]{0.07\textwidth}
            \captionsetup{font=scriptsize}
              \caption*{Original}\vspace{-2mm}
			\includegraphics[width=1\textwidth]{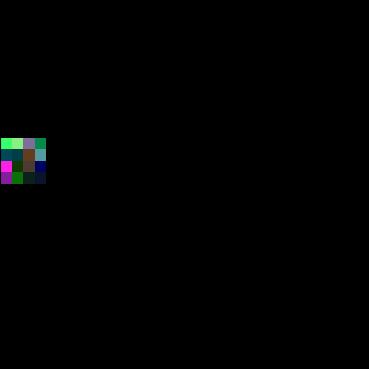} 
		\end{minipage}
	}\vspace{-2mm}
     \subfigure{
		\begin{minipage}[b]{0.07\textwidth}
              \captionsetup{font=scriptsize}
              \caption*{NC}\vspace{-2mm}
			\includegraphics[width=1\textwidth]{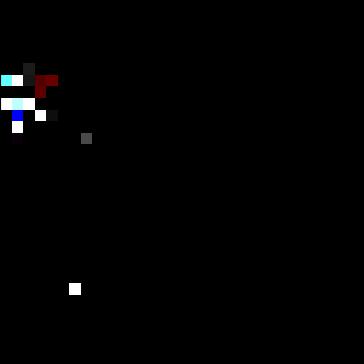}
		\end{minipage}
	}
     \subfigure{
		\begin{minipage}[b]{0.07\textwidth}
            \captionsetup{font=scriptsize}
            \caption*{TABOR}\vspace{-2mm}
			\includegraphics[width=1\textwidth]{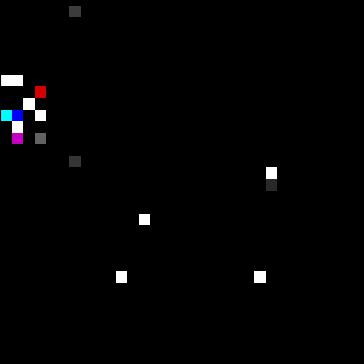}
		\end{minipage}
	}
         \subfigure{
		\begin{minipage}[b]{0.07\textwidth}
                \captionsetup{font=scriptsize}
                \caption*{USB}\vspace{-2mm}
			\includegraphics[width=1\textwidth]{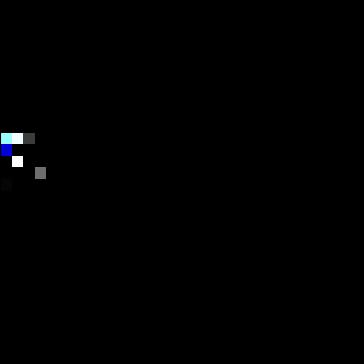}
		\end{minipage}
	}\\
        \subfigure{
        \rotatebox{90}{~~\scriptsize{ImageNet}}
		\begin{minipage}[b]{0.07\textwidth}
			\includegraphics[width=1\textwidth]{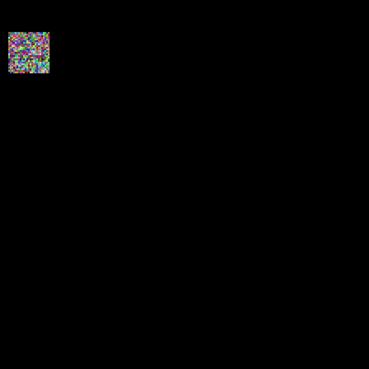} 
		\end{minipage}
	}\vspace{-2mm}
     \subfigure{
		\begin{minipage}[b]{0.07\textwidth}
			\includegraphics[width=1\textwidth]{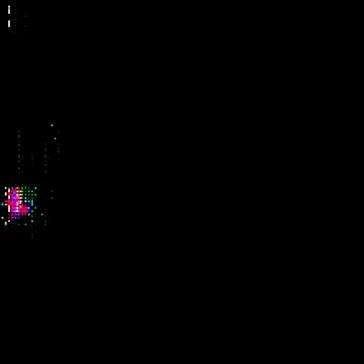}
		\end{minipage}
	}
     \subfigure{
		\begin{minipage}[b]{0.07\textwidth}
			\includegraphics[width=1\textwidth]{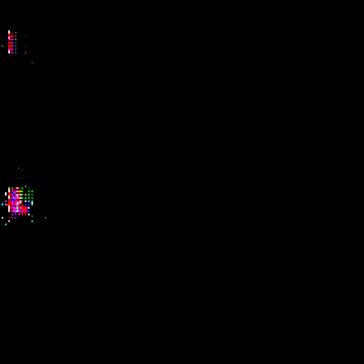}
		\end{minipage}
	}
         \subfigure{
		\begin{minipage}[b]{0.07\textwidth}
			\includegraphics[width=1\textwidth]{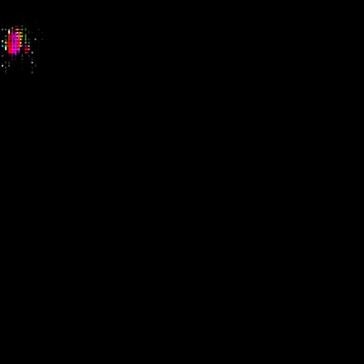}
		\end{minipage}
	}\\
         \subfigure{
         \rotatebox{90}{~~\scriptsize{ImageNet}}
		\begin{minipage}[b]{0.07\textwidth}
			\includegraphics[width=1\textwidth]{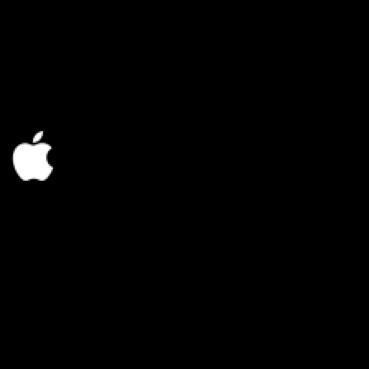} 
		\end{minipage}
	}
     \subfigure{
		\begin{minipage}[b]{0.07\textwidth}
			\includegraphics[width=1\textwidth]{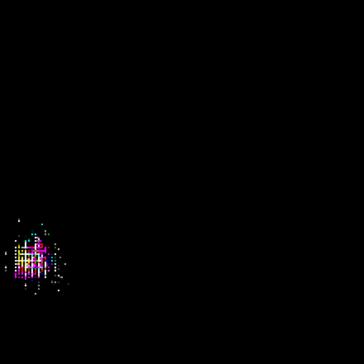}
		\end{minipage}
	}
     \subfigure{
		\begin{minipage}[b]{0.07\textwidth}
			\includegraphics[width=1\textwidth]{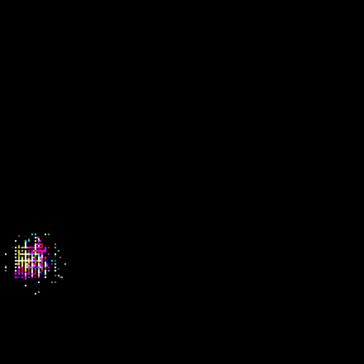}
		\end{minipage}
	}
         \subfigure{
		\begin{minipage}[b]{0.07\textwidth}
			\includegraphics[width=1\textwidth]{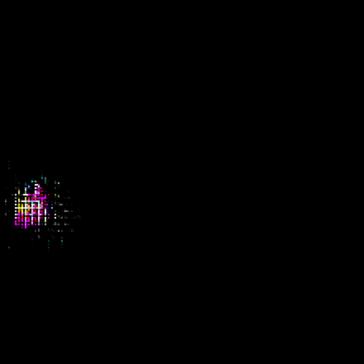}
		\end{minipage}
	}
	\caption{Examples of original triggers and reversed triggers by NC, TABOR, and \ourmethod for CIFAR-10 and ImageNet.}
	\label{fig:morereversetrigger}
\end{figure}

\subsection{Experimental Setup}

\noindent\textbf{Model, Datasets, and Backdoor.}
We use ResNet-18~\cite{He_2016_CVPR}, VGG-16~\cite{DBLP:journals/corr/SimonyanZ14a} on CIFAR-10~\cite{krizhevsky2009learning}, and Efficientnet-B0~\cite{pmlr-v97-tan19a} on ImageNet~\cite{5206848}.
We use BadNet~\cite{8685687}, Latent Backdoor~\cite{10.1145/3319535.3354209} and IAD attack~\cite{NEURIPS2020_234e6913} to inject backdoor into victim models.
The poisoning percent is 0.01.
The triggers are generated in different positions and random colors.

\noindent\textbf{Hyperparameters.}
To generate UAP, we use Hyperparameters following experiments in~\cite{Moosavi-Dezfooli_2017_CVPR}, and we set the desired error rate to $\theta=0.6$.
Alg.~\ref{alg:tuap} only relies on a small number, i.e., 300, of data points for $X$, while NC and TABOR use the entire training set as their input.
Then, the maximum iteration number of Alg.~\ref{alg:updateuap} is $m=500$, the learning rate (lr) is $lr=0.1$, and the optimizer is Adam (for detection) with $beta=(0.5, 0.9)$.
The hyperparameters to train clean and backdoored models are default ones from TrojanZoo~\cite{pang:2022:eurosp}:
batch size=96, lr=0.01, epoch=50 for CIFAR-10 and ImageNet.


\begin{figure}[t]
	\centering
     \subfigure[2$\times$2]{
		\begin{minipage}[b]{0.08\textwidth}
			\includegraphics[width=1\textwidth]{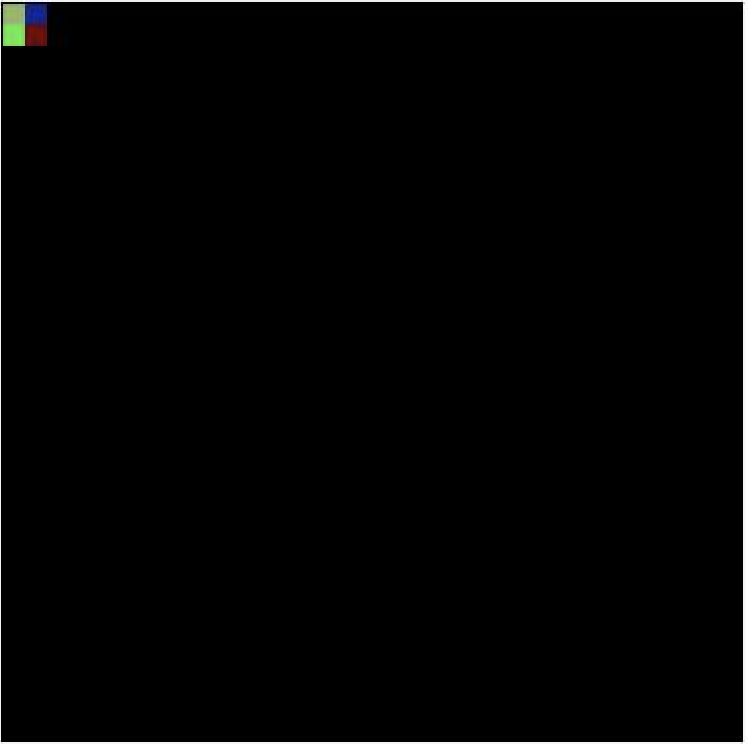} 
		\end{minipage}
		\label{fig:clafea22cifar10ori}
	}
     \subfigure[NC]{
		\begin{minipage}[b]{0.08\textwidth}
			\includegraphics[width=1\textwidth]{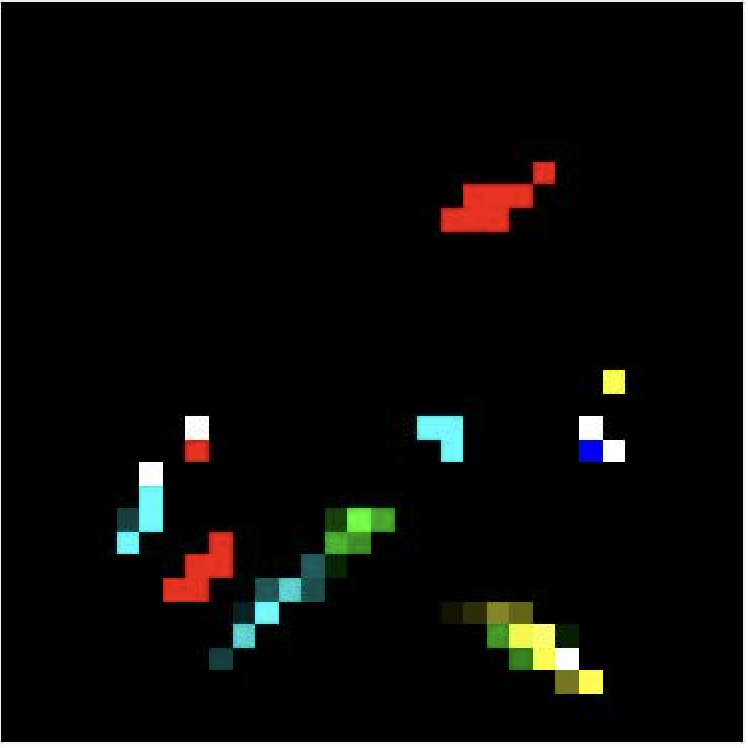}
		\end{minipage}
		\label{fig:clafea22cifar10nc}
	}
     \subfigure[TABOR]{
		\begin{minipage}[b]{0.08\textwidth}
			\includegraphics[width=1\textwidth]{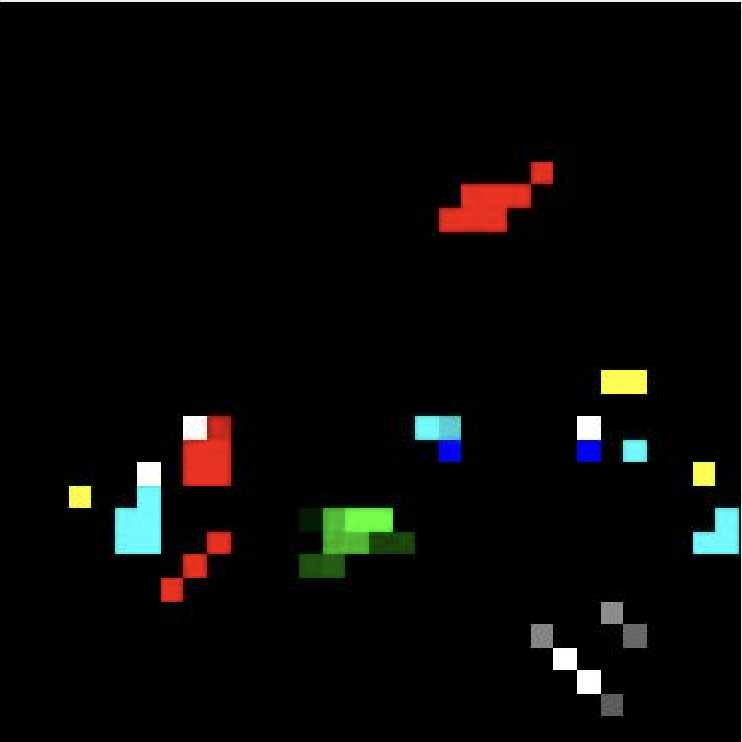}
		\end{minipage}
		\label{fig:clafea22cifar10tabor}
	}
         \subfigure[\ourmethod]{
		\begin{minipage}[b]{0.08\textwidth}
			\includegraphics[width=1\textwidth]{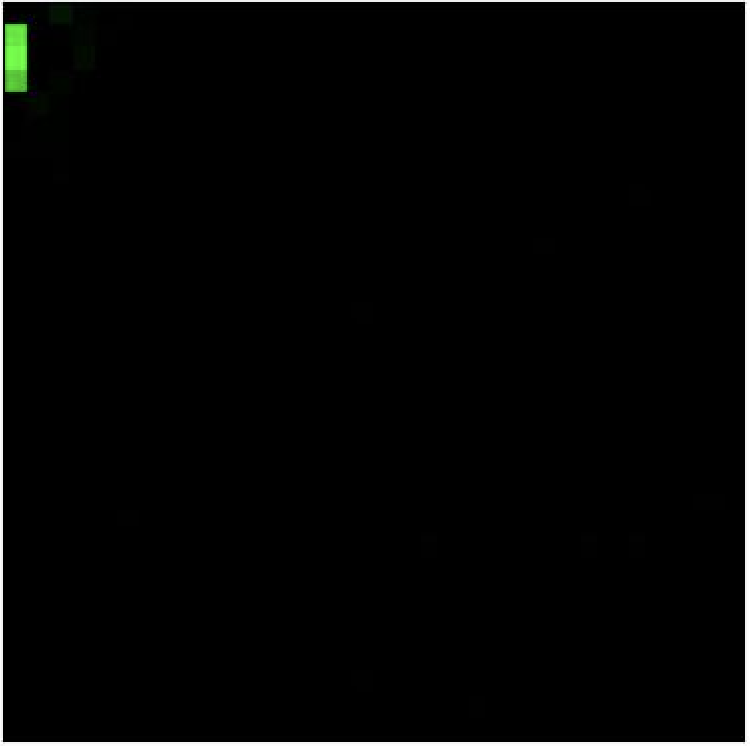}
		\end{minipage}
		\label{fig:clafea22cifar10our}
	}
	\caption{An example visualization of the original trigger and reversed triggers by NC, TABOR, and \ourmethod for CIFAR-10.}
	\label{fig:classfeaturetriggers}
\end{figure}

\noindent\textbf{Evaluation.}
Following the previous work in~\cite{Dong_2021_ICCV},
we design two metrics to evaluate the defense performance: model detection and target class detection.
We check whether a model is correctly identified as a clean or backdoored model.
Then, for backdoored models, we check whether reverse engineering correctly identifies the target class.
In Tab.~\ref{tab:cifar10detection},  Tab.~\ref{tab:imagenet_efficientnet_detection}, and Tab.~\ref{tab:cifarvgg16otherattacks}, \textit{Clean} and \textit{Backdoored} under \textit{Model Detection} refer to the cases whether the detection identifies a model as clean or backdoored.
For \textit{Target Class Detection}, we have three categories: (i) \textit{Correct} means the detection method identifies the true target class of a backdoored model, (ii) \textit{Correct Set} refers to the case where the detection method identifies multiple backdoors on different classes, including the true target class, and (iii) \textit{Wrong} refers to the case where the detection method successfully identifies a backdoored model but with wrong target class(es).

\begin{table*}[tb]
\footnotesize
\centering
\setlength\tabcolsep{2pt}
\begin{tabular}{cccccccccccccc}
\toprule
\multirow{2}{*}{Model} & \multirow{2}{*}{Accuracy} & \multirow{2}{*}{ASR} & \multirow{2}{*}{Method} & \multicolumn{1}{c}{Reversed Trigger} & \multicolumn{2}{c}{Model Detection} & \multicolumn{3}{c}{Target Class Detection} \\
~& ~& ~& ~& $L_1$ norm & Clean & Backdoored & Correct  & Correct Set & Wrong\\
\midrule
\multirow{3}{*}{\tabincell{c}{Backdoored \\ (20 $\times$ 20 trigger)}} & \multirow{3}{*}{70.94} & \multirow{3}{*}{76.67} & NC & 276.78 & 0  & 15  & 14 & 1 & 0\\
~& ~& ~& TABOR& 271.83 & 0  & 15  & 12 & 2 & 1\\
~& ~& ~& \ourmethod& 461.32 & 0  & 15  & 14  & 1  & 0 \\
\midrule
\multirow{3}{*}{\tabincell{c}{Backdoored \\ (25 $\times$ 25 trigger)}} & \multirow{3}{*}{69.7} & \multirow{3}{*}{78.46} & NC & 347.48 & 0  & 15  & 13 & 2 & 0 \\
~& ~& ~& TABOR& 341.47 & 2 & 13  & 13 & 0 & 0\\
~& ~& ~& \ourmethod& 547.56 & 0  & 15  & 15  & 0  & 0   \\
\midrule
\multirow{3}{*}{\begin{minipage}[b]{0.06\columnwidth}
		\centering
		\raisebox{-.5\height}{\includegraphics[width=\linewidth]{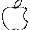}}
	\end{minipage}} & \multirow{3}{*}{70.91} & \multirow{3}{*}{80.02} & NC & 396.72 & 1  & 14  & 14 & 0 & 0\\
~& ~& ~& TABOR& 406.1 & 3 & 12  & 12 & 0 & 0\\
~& ~& ~& \ourmethod& 621.0 & 1  & 14  & 14  & 0  & 0  \\
\bottomrule
\end{tabular}
\caption{Detection evaluation on ImageNet where each case consists of 15 trained models.}
\label{tab:imagenet_efficientnet_detection}
\end{table*}

\begin{table*}[tb]
\footnotesize
\centering
\setlength\tabcolsep{2pt}
\begin{tabular}{cccccccccccccc}
\toprule
\multirow{2}{*}{Model} & \multirow{2}{*}{Accuracy} & \multirow{2}{*}{ASR} & \multirow{2}{*}{Method} & \multicolumn{1}{c}{Reversed Trigger} & \multicolumn{2}{c}{Model Detection} & \multicolumn{3}{c}{Target Class Detection} \\
~& ~& ~& ~& $L_1$ norm & Clean & Backdoored & Correct  & Correct Set & Wrong\\
\midrule
 \multirow{3}{*}{Clean} & \multirow{3}{*}{91.59} & \multirow{3}{*}{N/A} & NC & 40.78 & 15 & 0 & N/A & N/A & N/A\\
~& ~& ~& TABOR& 48.53 & 14& 1 & 0 & 0 & 1\\
~& ~& ~& \ourmethod& 47.53 & 15 & 0 & N/A & N/A & N/A\\
\midrule
\multirow{3}{*}{\tabincell{c}{Latent Backdoor \\ (4 $\times$ 4 trigger)}} & \multirow{3}{*}{87.20} & \multirow{3}{*}{99.66} & NC & 19.71 & 4 & 11 & 10 & 1 & 0 \\
~& ~& ~& TABOR& 20.68 & 4 & 11 & 11 & 0 & 0\\
~& ~& ~& \ourmethod& 12.37 & 1 & 14 & 13 & 1 & 0  \\
\midrule
\multirow{3}{*}{\tabincell{c}{Input Aware\\Dynamic\\ (32 $\times$ 32 trigger)}} & \multirow{3}{*}{89.46} & \multirow{3}{*}{90.43} & NC & 0.0 & 15 & 0 & N/A & N/A & N/A\\
~& ~& ~& TABOR& 1.8 & 15 & 0 & N/A & N/A & N/A\\
~& ~& ~& \ourmethod& 0.13 & 0 & 15 & 15 & 0 & 0\\
\bottomrule
\end{tabular}
\caption{Detection evaluation by stronger backdoor attacks on VGG-16 trained with CIFAR-10.}
\label{tab:cifarvgg16otherattacks}
\end{table*}

\subsection{Results}

Tab.~\ref{tab:cifar10detection} shows the detection results for CIFAR-10.
In CIFAR-10, class may contain features from other classes.
For example, ``cat'' and ``dog'' share the feature of four limbs.
For the backdoored models, \ourmethod achieves a higher accuracy (98\%) on detecting backdoored models compared to NC (93\%) and TABOR (92\%).
We believe that the misclassifications in NC and TABOR are caused by capturing the class's unique features rather than the trigger, illustrated in Fig.~\ref{fig:classfeaturetriggers}. 

As ImageNet contains a vast number of images, it is hard to train a lot of models on it. 
Thus we use a subset of ImageNet, which contains ten classes. 
Each class has 1301 images. 
Tab.~\ref{tab:imagenet_efficientnet_detection} shows the results of detecting backdoors for Efficientnet-B0~\cite{pmlr-v97-tan19a} trained with the subset of ImageNet~\cite{5206848}.
Due to the larger image size and model architecture, we use 500 images for data points $X$ in Alg.~\ref{alg:tuap} and Alg.~\ref{alg:updateuap}.

\subsection{Stronger Backdoor Attacks}
Tab.~\ref{tab:cifarvgg16otherattacks} shows the detection results on Latent Backdoor~\cite{10.1145/3319535.3354209} and IAD attack~\cite{NEURIPS2020_234e6913}. 
The trigger size for Latent Backdoor is $4 \times 4 \times 3$.
Due to IAD attack's characteristics, we use $32 \times 32 \times 3$ trigger size (the size of input images).
The motivation is to show the generalization of \ourmethod under stronger attacks besides the BadNets~\cite{8685687}, especially IAD is non-patch-wised.
IAD trigger changes with different inputs.

According to Tab.~\ref{tab:cifarvgg16otherattacks}, NC and TABOR show worse performance compared to detection results on BadNets,
while \ourmethod still precisely detects most of the backdoored models.
Especially, NC and TABOR do not work under the IAD attack, but \ourmethod detects such backdoors with the true target class. 
The reason is that NC-style methods do not directly optimize the \textbf{Trigger patterns}. 
They mainly optimize the mask that will be applied to the pattern.
Moreover, IAD attacks designs subtle triggers with specific features related to inputs, which is more difficult for optimization from random points.

\subsection{Time Consumption}
As mentioned before, NC and TABOR require a large amount iterations to conduct detection.
Therefore, we evaluate the time consumption of NC, TABOR and \ourmethod when conducting detection with Efficientnet-B0 on ImageNet.
When detecting backdoored models with $20 \times 20$ trigger, the average time consumption (in second) for NC, TABOR and \ourmethod are: 1154.02, 2.129.40 and 267.12, respectively.
It is clear that \ourmethod costs much less time when optimizing reverse engineering the potential triggers.
Although \ourmethod needs to generate targeted UAP, the UAP can be used for different models with similar architecture~\cite{Moosavi-Dezfooli_2017_CVPR}.
We only need to generate it once.



\section{Conclusions and Future Work}
\label{sec:conclusions}

This paper proposes \ourmethod to detect potential backdoors.
\ourmethod uses targeted UAP to capture sensitive features created by backdoors.
Then the UAP is optimized to focus on pixels sensitive to backdoors.
The aim is to avoid using the class features as the backdoor triggers.
We run extensive experiments on several datasets to evaluate the performance of our method.
Among the 160 backdoored models, we successfully identified 157 backdoored ones and outperform baselines.
Finally, further investigation is needed regarding optimizing UAP according to backdoored neurons in the backddoored model, which can greatly reduce the optimization time.


\bibliographystyle{IEEEbib}
\bibliography{uap}

\clearpage

 \appendix
 \section{Appendix}
\subsection{Experimental Results on More Settings}
We demonstrate the generalization of \ourmethod by considering more settings in Appendix:
\begin{itemize}
    \item Stronger backdoor attacks. We study the Latent Backdoor~\cite{10.1145/3319535.3354209} and Input-Aware Dynamic Backdoor Attack~\cite{NEURIPS2020_234e6913} besides BadNets~\cite{8685687}.
    \item Different model architectures. We study the VGG-16~\cite{DBLP:journals/corr/SimonyanZ14a} besides ResNet-18~\cite{He_2016_CVPR} and Efficientnet-B0~\cite{pmlr-v97-tan19a}.
    \item More datasets. We also include detection results with GTSRB~\cite{Stallkamp2012} besides MNIST~\cite{lecun1998mnist}, CIFAR-10~\cite{krizhevsky2009learning}, and ImageNet~\cite{5206848}.
\end{itemize}

\begin{table*}[tb]
\footnotesize
\centering
\setlength\tabcolsep{2pt}
\begin{tabular}{cccccccccccccc}
\toprule
\multirow{2}{*}{Model} & \multirow{2}{*}{Accuracy} & \multirow{2}{*}{ASR} & \multirow{2}{*}{Method} & \multicolumn{1}{c}{Reversed Trigger} & \multicolumn{2}{c}{Model Detection} & \multicolumn{3}{c}{Target Class Detection} \\
~& ~& ~& ~& $L_1$ norm & Clean & Backdoored & Correct  & Correct Set & Wrong\\
\midrule
 \multirow{3}{*}{Clean} & \multirow{3}{*}{91.59} & \multirow{3}{*}{N/A} & NC & 40.78 & 15 & 0 & N/A & N/A & N/A\\
~& ~& ~& TABOR& 48.53 & 14& 1 & 0 & 0 & 1\\
~& ~& ~& \ourmethod& 47.53 & 15 & 0 & N/A & N/A & N/A\\
\midrule
\multirow{3}{*}{\tabincell{c}{Backdoored \\ (2 $\times$ 2 trigger)}} & \multirow{3}{*}{88.28} & \multirow{3}{*}{99.39} & NC & 5.43 & 0 & 15 & 14 & 1 &0\\
~& ~& ~& TABOR& 5.32 & 0& 15 & 15 & 0 & 0\\
~& ~& ~& \ourmethod& 3.5 & 0 & 15 & 15 & 0 & 0 \\
\midrule
\multirow{3}{*}{\tabincell{c}{Backdoored \\ (3 $\times$ 3 trigger)}} & \multirow{3}{*}{88.30} & \multirow{3}{*}{99.77} & NC & 6.60 & 1 & 14 & 13 & 1 & 0 \\
~& ~& ~& TABOR& 6.98 & 0 & 15 & 14 & 1 & 0\\
~& ~& ~& \ourmethod& 7.0 & 0 & 15 & 14 & 1 & 0  \\
\bottomrule
\end{tabular}
\caption{Detection evaluation on VGG-16 trained with CIFAR-10 where each case consists of 15 trained models.}
\label{tab:cifarvgg16detection}
\end{table*}

\begin{figure}[t]
	\centering
	\subfigure[$2\times2$]{
		\begin{minipage}[b]{0.1\textwidth}
			\includegraphics[width=1\textwidth]{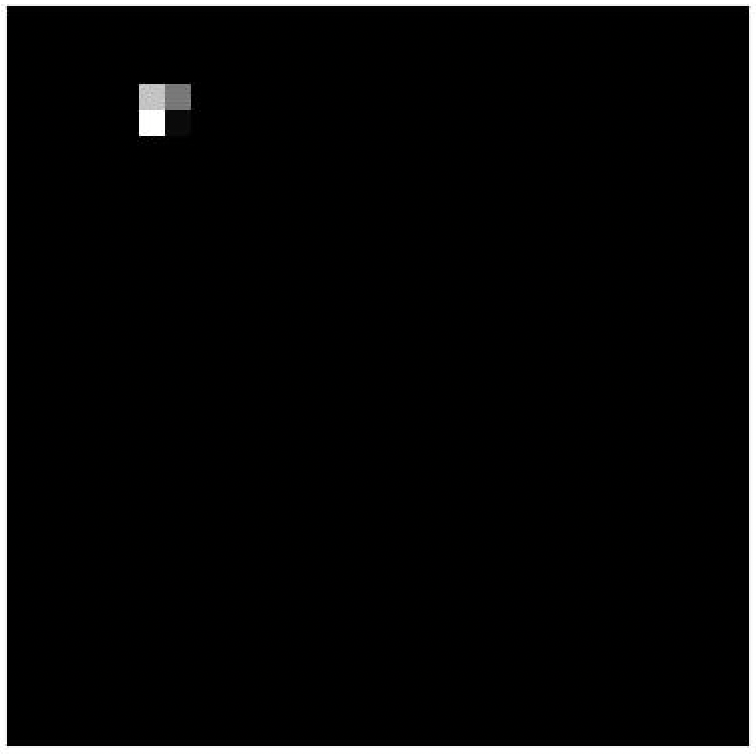} 
		\end{minipage}
		\label{fig:22ori}
	}
    	\subfigure[NC]{
    		\begin{minipage}[b]{0.1\textwidth}
   		 	\includegraphics[width=1\textwidth]{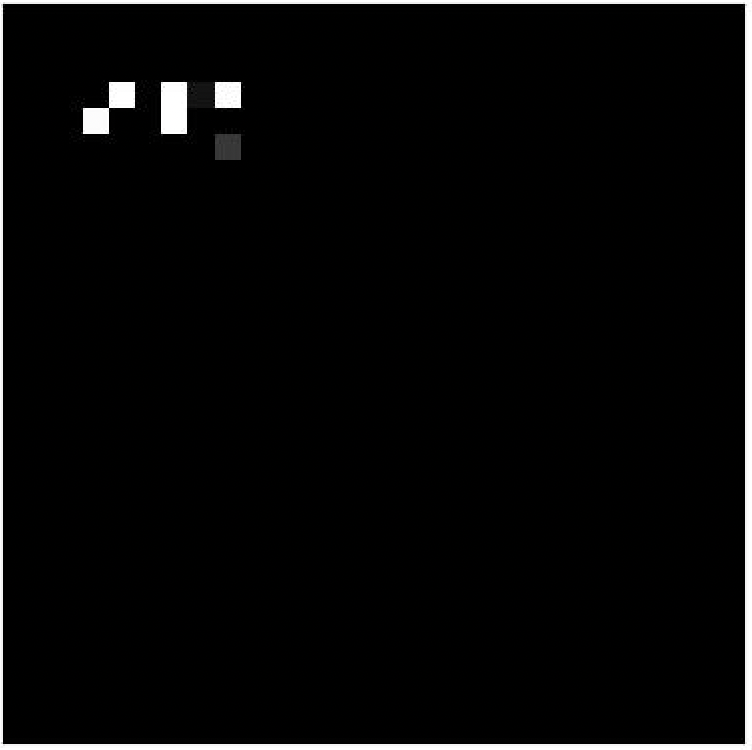}
    		\end{minipage}
		\label{fig:22nc}
    	}
	\subfigure[TABOR]{
		\begin{minipage}[b]{0.1\textwidth}
			\includegraphics[width=1\textwidth]{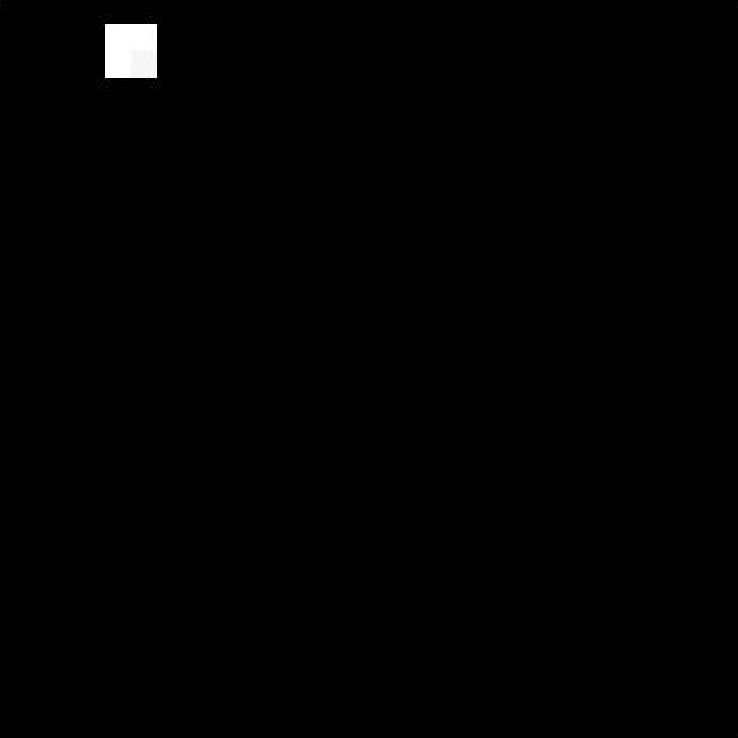} 
		\end{minipage}
		\label{fig:22tabor}
	}
         \subfigure[\ourmethod]{
		\begin{minipage}[b]{0.1\textwidth}
			\includegraphics[width=1\textwidth]{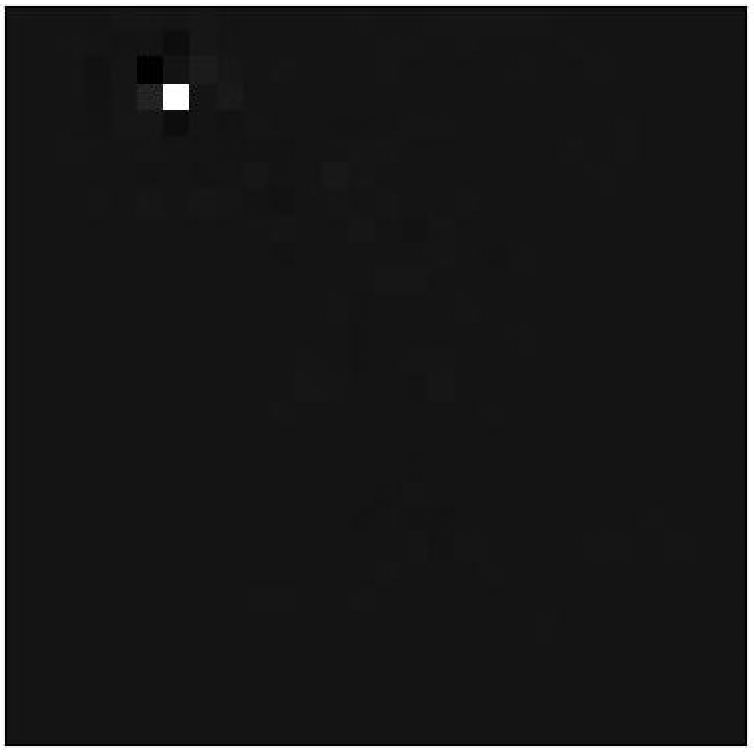} 
		\end{minipage}
		\label{fig:22our}
	}
	\subfigure[$3\times3$]{
		\begin{minipage}[b]{0.1\textwidth}
			\includegraphics[width=1\textwidth]{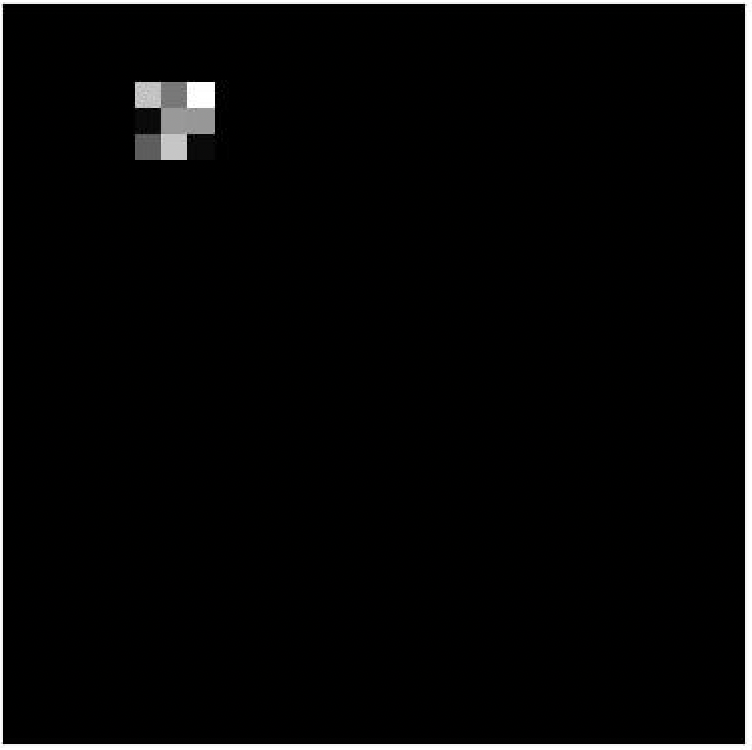} 
		\end{minipage}
		\label{fig:33ori}
	}
    	\subfigure[NC]{
    		\begin{minipage}[b]{0.1\textwidth}
   		 	\includegraphics[width=1\textwidth]{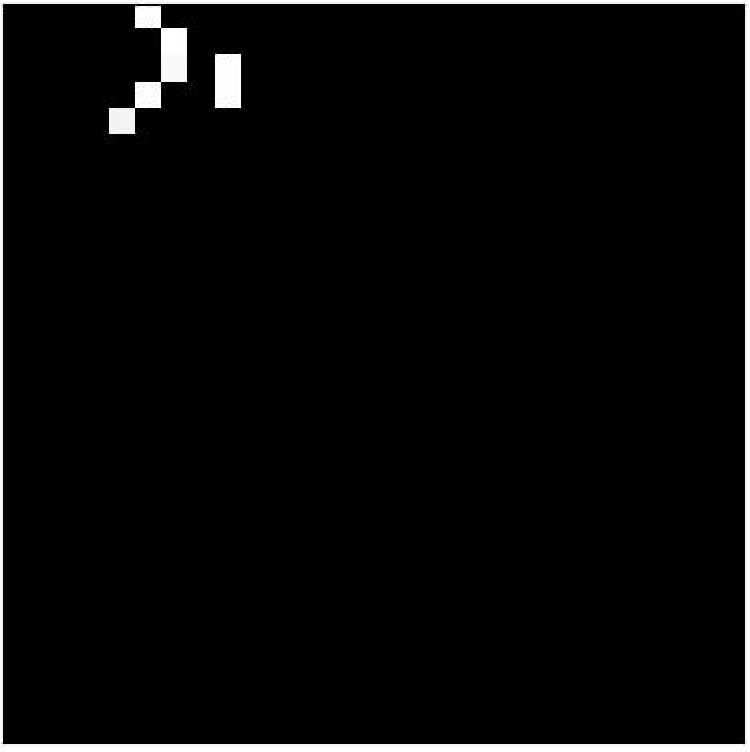}
    		\end{minipage}
		\label{fig:33nc}
    	}
     \subfigure[TABOR]{
		\begin{minipage}[b]{0.1\textwidth}
			\includegraphics[width=1\textwidth]{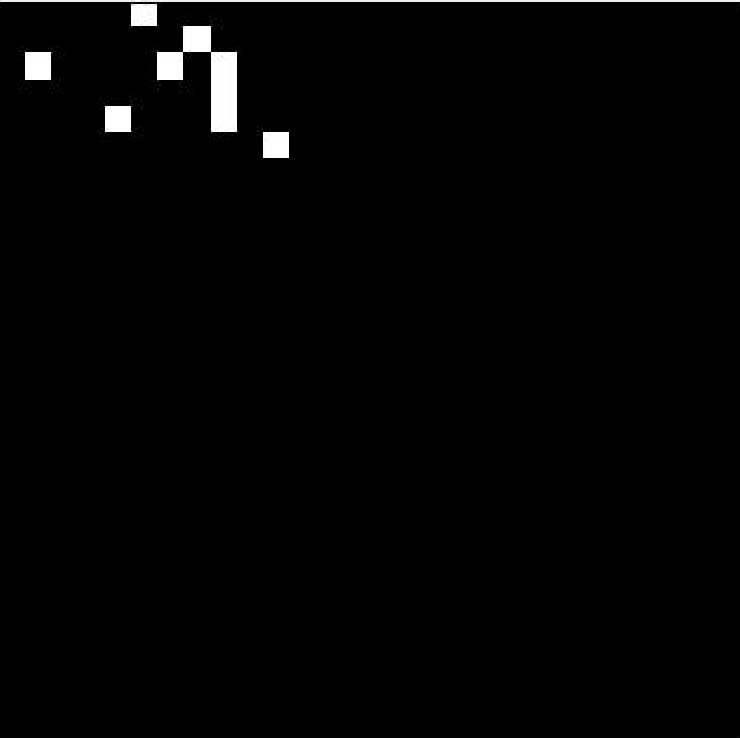} 
		\end{minipage}
		\label{fig:33tabor}
	}
	\subfigure[\ourmethod]{
		\begin{minipage}[b]{0.1\textwidth}
			\includegraphics[width=1\textwidth]{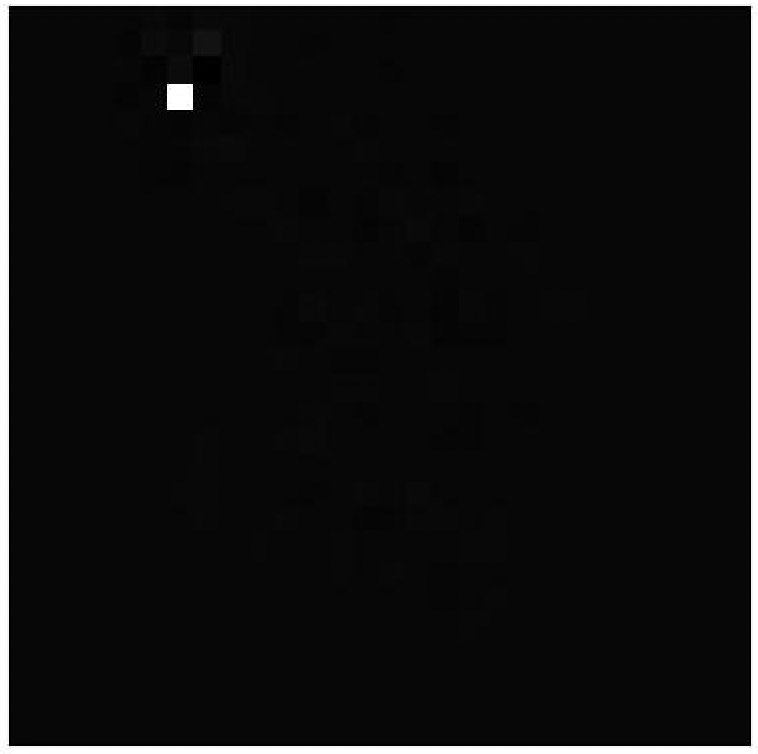} 
		\end{minipage}
		\label{fig:33our}
	}
	\caption{An example visualization of the original triggers and reversed triggers. Fig.~\ref{fig:22ori} and Fig.~\ref{fig:33ori} are original triggers and followed by three reversed triggers. More examples in Appendix.}
	\label{fig:triggers}
\end{figure}

\subsection{MNIST}\label{sec:experimentmnist}
We compare \ourmethod with NC and TABOR while they use the whole clean training set to conduct optimization. 
Tab.~\ref{tab:mnistdetection} shows detection results, including average accuracy on clean data, attack success rates (ASRs), average $L_1$ norm of the reversed trigger, and detection evaluation.
According to detection results, neither our method nor NC and TABOR mistake clean models for backdoor models.
The $L_1$ norm values of triggers generated from clean models for every class are close to each other.
There are no outliers among these norm values.
On backdoored models, all three methods can identify the majority of backdoors.
In the wrong cases, \ourmethod can still detect backdoors, although the target class is not the true target.
NC and TABOR tend to classify wrong cases as clean models.

\begin{table*}[ht]
\footnotesize
\centering
\setlength\tabcolsep{2pt}
\begin{tabular}{cccccccccccccc}
\toprule
\multirow{2}{*}{Model} & \multirow{2}{*}{Accuracy} & \multirow{2}{*}{ASR} & \multirow{2}{*}{Method} & \multicolumn{1}{c}{Reversed Trigger} & \multicolumn{2}{c}{Model Detection} & \multicolumn{3}{c}{Target Class Detection} \\
~& ~& ~& ~& $L_1$ norm & Clean & Backdoored & Correct  & Correct Set & Wrong\\
\midrule
 \multirow{3}{*}{Clean} & \multirow{3}{*}{98.93} & \multirow{3}{*}{N/A} & NC & 91.50 & 50 & 0 & N/A & N/A & N/A\\
~& ~& ~& TABOR& 95.31 & 50& 0 & N/A & N/A & N/A\\
~& ~& ~& \ourmethod& 44.23 & 50 & 0 & N/A & N/A & N/A\\
\midrule
\multirow{3}{*}{\tabincell{c}{Backdoored \\ (2 $\times$ 2 trigger)}} & \multirow{3}{*}{98.82} & \multirow{3}{*}{94.43} & NC & 6.37 & 1 & 49 & 49 & 0 &0\\
~& ~& ~& TABOR& 6.57 & 1& 49 & 49 & 0 & 0\\
~& ~& ~& \ourmethod& 9.22 & 0 & 50 & 49 & 0 & 1 \\
\midrule
\multirow{3}{*}{\tabincell{c}{Backdoored \\ (3 $\times$ 3 trigger)}} & \multirow{3}{*}{99.00} & \multirow{3}{*}{99.53} & NC & 7.89 & 1 & 49 & 49 & 0 & 0 \\
~& ~& ~& TABOR& 6.04 & 0 & 50 & 50 & 0 & 0\\
~& ~& ~& \ourmethod& 8.11 & 0 & 50 & 49 & 1 & 0  \\
\bottomrule
\end{tabular}
\caption{Detection evaluation on MNIST where each case consists of 50 trained models.}
\label{tab:mnistdetection}
\end{table*}

\subsection{Detection Results on VGG-16}
In Tab.~\ref{tab:cifarvgg16detection}, we show the results of detecting backdoors for VGG-16 models trained with CIFAR-10.
We use the same experimental settings as that in the experiment section.
We also study Latent Backdoor~\cite{10.1145/3319535.3354209} beside BadNet attack. 

\subsection{Stronger Backdoor Attacks}
The basic idea of the NC-style method is optimizing the \textbf{Mask} so that the final trigger (\textbf{Trigger pattern} $\times$ \textbf{Mask}) can mislead the classifier to the target class.
The optimization starts from a random point.
However, NC-style methods do not directly optimize the \textbf{Trigger patterns}. 
In Fig.~\ref{fig:startpoints}, it is clear that the optimized trigger pattern is almost the same as the random start point.
Moreover, the optimization process can be more difficult to generate correct triggers when the random start point is far away from optimum.
\ourmethod uses similar optimization to find potential triggers, but we separate the process into two different parts.
In the first part (Section~\ref{sec:method}), we search for a targeted UAP as the starting point for reverse engineering of potential triggers.
The UAP already includes latent features of the model to be detected, while a random start point includes lots of noise, see Fig.~\ref{fig:startpoints} as an example.
In other words, if the original trigger is not included in the random start, it will be difficult for NC-style methods to be successful.
Stronger attacks, such as Input-Aware attacks, design subtle triggers with specific features related to inputs.
Random starting points cannot include these specific features.
Therefore, NC and TABOR cannot detect the backdoor under an Input-Aware attack.

\subsection{GTSRB}
The results for GTSRB are shown in Tab.~\ref{tab:GTSRBdetection}.
On clean models, \ourmethod, NC, and TABOR all have incorrect results, as the number of classes in GTSRB is significantly larger than that of MNIST and CIFAR-10.
Compared to the $L_1$ norm of NC and TABOR, \ourmethod provides a much smaller norm value because the reversed trigger is optimized from targeted UAP.
Since GTSRB has more classes than MNIST and CIFAR-10, initialization with targeted UAP helps the trigger optimization (Alg.~\ref{alg:updateuap}) to avoid local optima and find as small as possible triggers.
However, \ourmethod generates more wrong cases than NC and TABOR.
The reason is that \ourmethod uses only 300 entries of clean data to conduct reverse engineering, which is to be consistent with experiments on MNIST and CIFAR-10.
On MNIST and CIFAR-10, there are around 30 entries for each class, but less than 10 entries for each class of GTSRB.
Such a small amount of data cannot provide enough features and information.
This gives a simple option to improve the performance of our method by adding more data.

\begin{table*}[ht]
\footnotesize
\centering
\setlength\tabcolsep{2pt}
\begin{tabular}{ccccccccccc}
\toprule
\multirow{2}{*}{Model} & \multirow{2}{*}{Accuracy} & \multirow{2}{*}{ASR} & \multirow{2}{*}{Method} & \multicolumn{1}{c}{Reversed Trigger} & \multicolumn{2}{c}{Model Detection} & \multicolumn{3}{c}{Target Class Detection} \\
~& ~& ~& ~& $L_1$ norm & Clean & Backdoored & Correct  & Correct Set & Wrong\\
\midrule
\multirow{3}{*}{Clean} & \multirow{3}{*}{83.96} & \multirow{3}{*}{N/A} & NC & 181.17 & 12  & 3  & N/A & N/A & N/A\\
~& ~& ~& TABOR& 185.21 & 13  & 2  & N/A & N/A & N/A\\
~& ~& ~& \ourmethod& 39.8 & 12  & 3  & N/A & N/A & N/A\\
\midrule
\multirow{3}{*}{\tabincell{c}{Backdoored \\ (2 $\times$ 2 trigger)}} & \multirow{3}{*}{80.85} & \multirow{3}{*}{85.06} & NC & 13.36 & 0  & 15  & 13  & 2  & 0  \\
~& ~& ~& TABOR& 37.02 & 0  & 15  & 13  & 2  & 0  \\
~& ~& ~& \ourmethod& 10.86 & 3  & 12  & 12 & 0 & 0 \\
\midrule
\multirow{3}{*}{\tabincell{c}{Backdoored \\ (3 $\times$ 3 trigger)}} & \multirow{3}{*}{80.24} & \multirow{3}{*}{93.52} & NC & 14.78 & 0  & 15  & 13  & 2  & 0 \\
~& ~& ~& TABOR& 15.11 & 0  & 15  & 13  & 2  & 0 \\
~& ~& ~& \ourmethod& 12.02 & 2  & 13  & 13 & 0 & 0 \\
\bottomrule
\end{tabular}
\caption{Detection evaluation on GTSRB where each case consists of  15 trained models.}
\label{tab:GTSRBdetection}
\end{table*}

\begin{figure}[h]
\centering
\subfigure{
		\begin{minipage}[b]{0.065\textwidth}
			\includegraphics[width=1\textwidth]{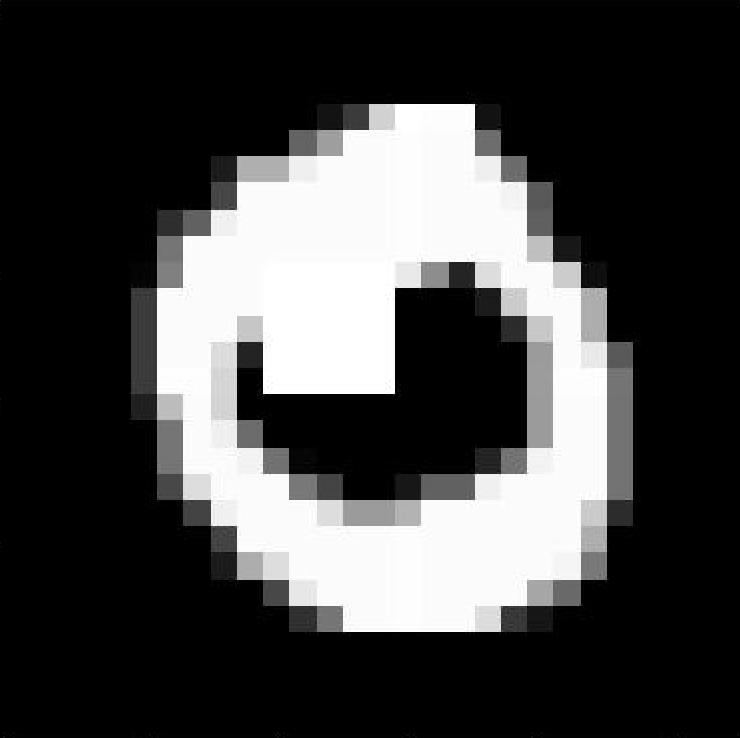} 
		\end{minipage}
		\label{fig:reverse_results_ori}
	}
\subfigure{
		\begin{minipage}[b]{0.065\textwidth}
			\includegraphics[width=1\textwidth]{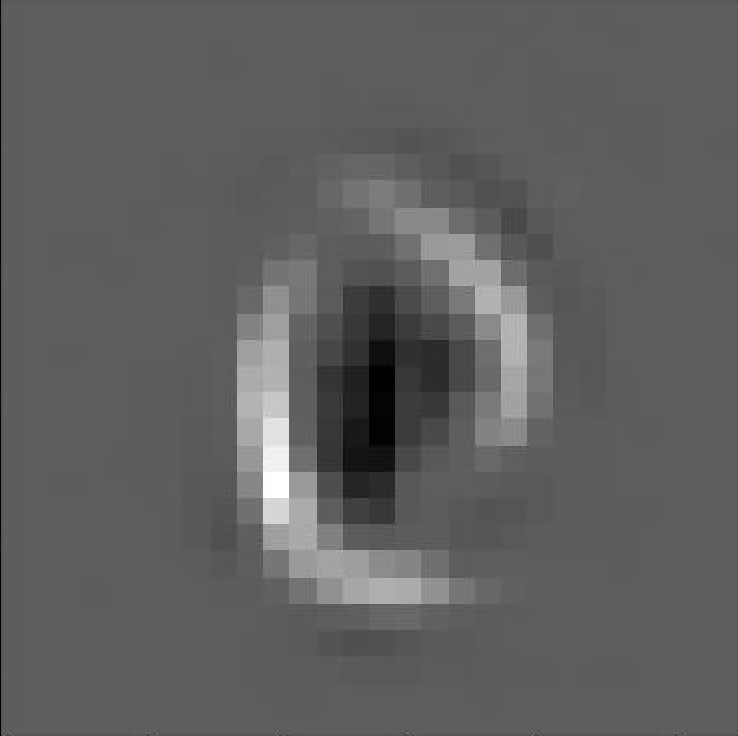} 
		\end{minipage}
		\label{fig:reverse_results_0}
	}
\subfigure{
		\begin{minipage}[b]{0.065\textwidth}
			\includegraphics[width=1\textwidth]{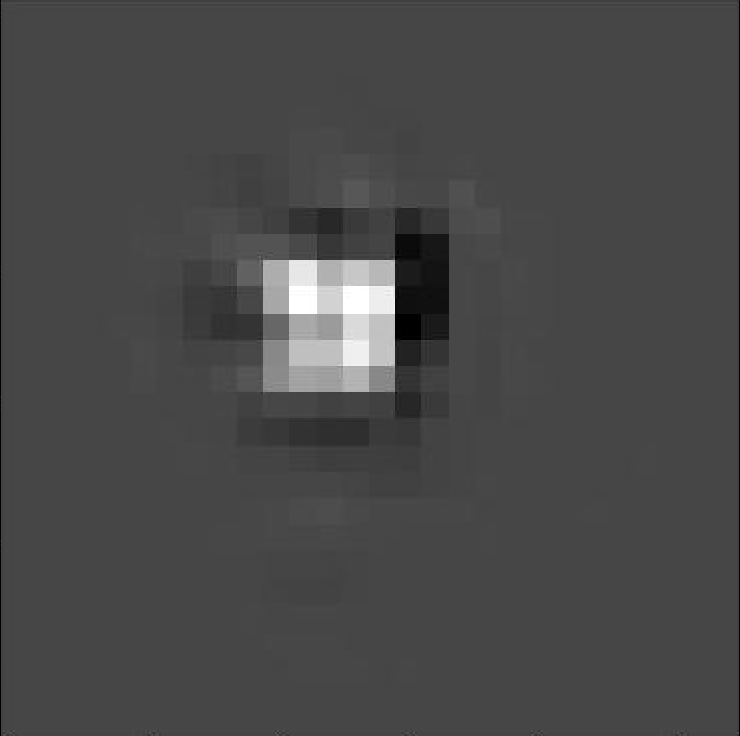} 
		\end{minipage}
		\label{fig:reverse_results_1}
	}
\subfigure{
		\begin{minipage}[b]{0.065\textwidth}
			\includegraphics[width=1\textwidth]{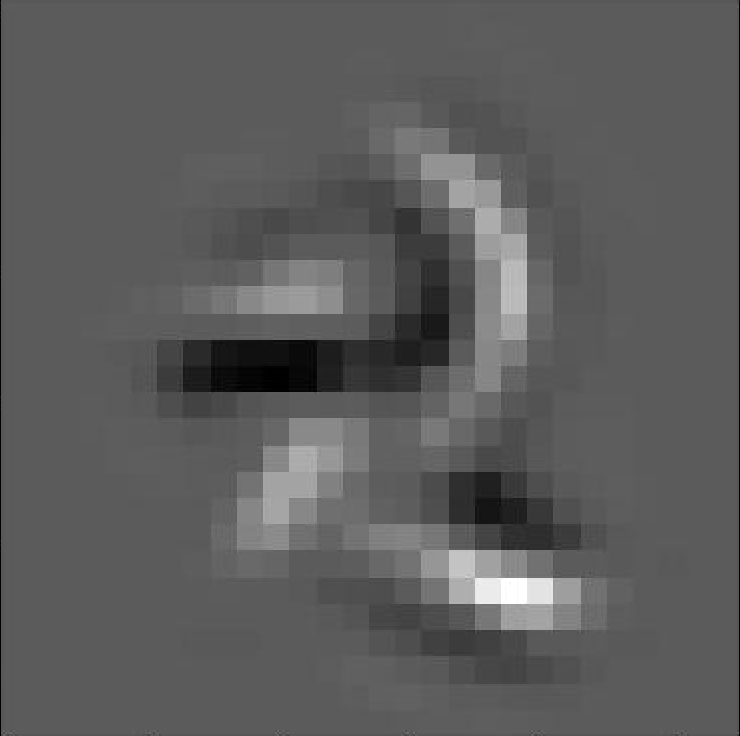} 
		\end{minipage}
		\label{fig:reverse_results_2}
	}
\subfigure{
		\begin{minipage}[b]{0.065\textwidth}
			\includegraphics[width=1\textwidth]{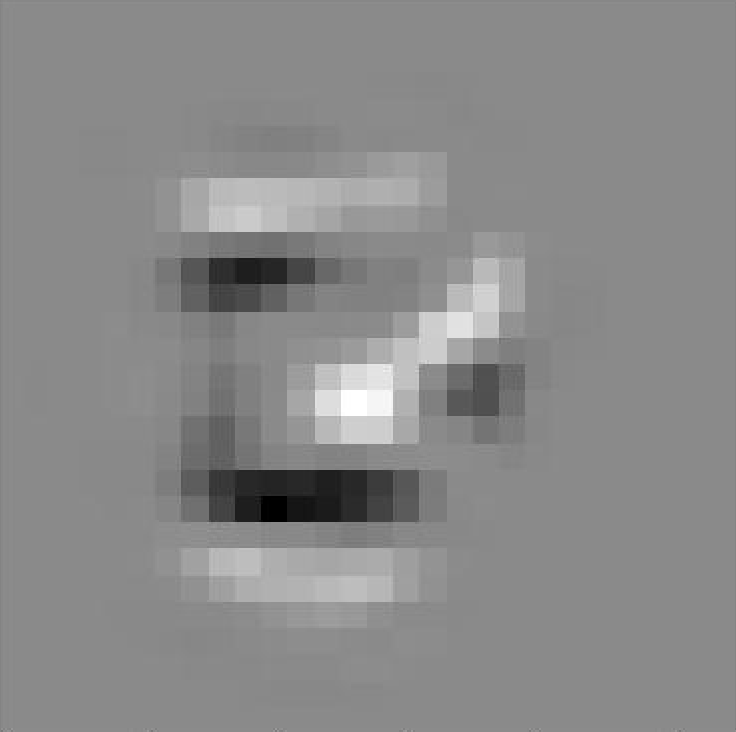} 
		\end{minipage}
		\label{fig:reverse_results_3}
	}
\subfigure{
		\begin{minipage}[b]{0.065\textwidth}
			\includegraphics[width=1\textwidth]{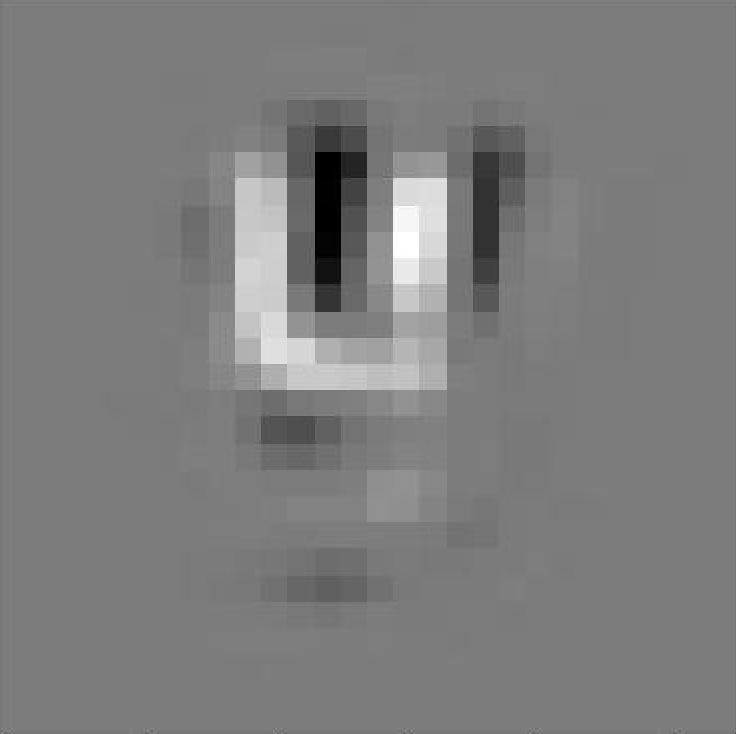} 
		\end{minipage}
		\label{fig:reverse_results_4}
	}
\subfigure{
		\begin{minipage}[b]{0.065\textwidth}
			\includegraphics[width=1\textwidth]{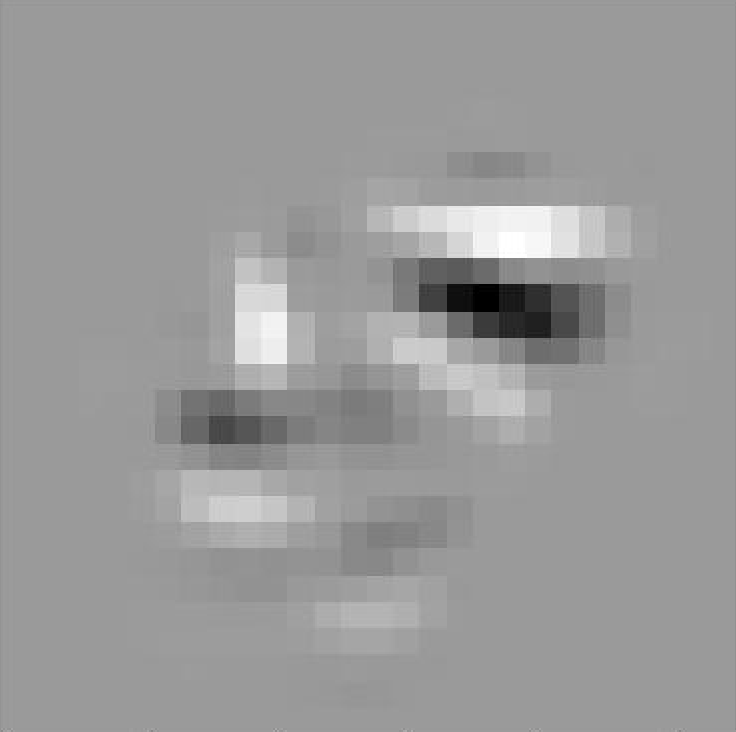} 
		\end{minipage}
		\label{fig:reverse_results_5}
	}
	\subfigure{
		\begin{minipage}[b]{0.065\textwidth}
			\includegraphics[width=1\textwidth]{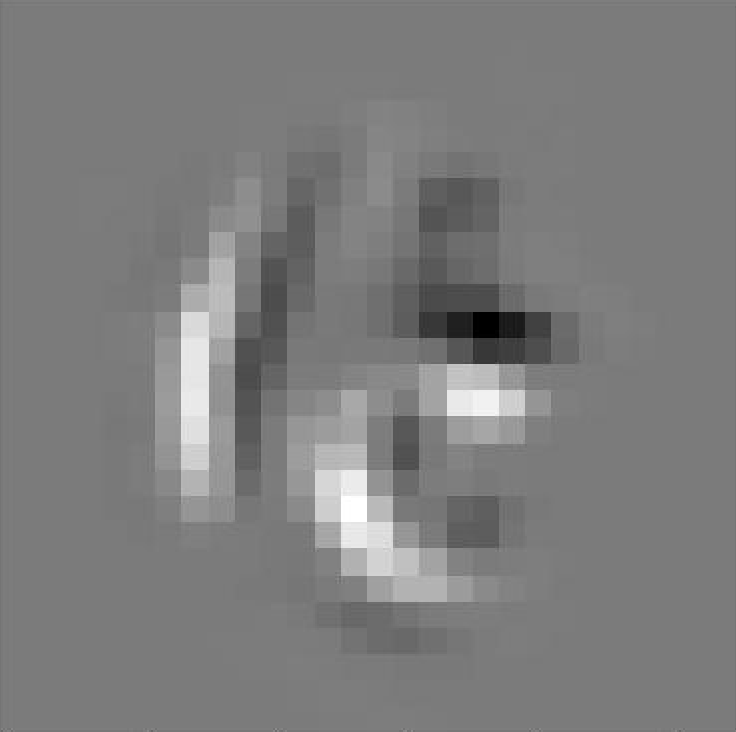} 
		\end{minipage}
		\label{fig:reverse_results_6}
	}
	\subfigure{
		\begin{minipage}[b]{0.065\textwidth}
			\includegraphics[width=1\textwidth]{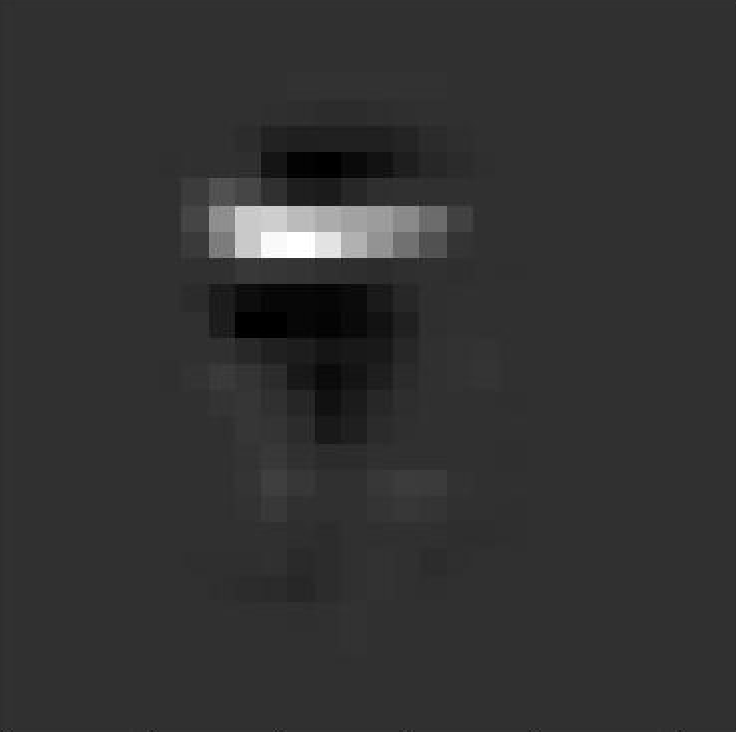} 
		\end{minipage}
		\label{fig:reverse_results_7}
	}
	\subfigure{
		\begin{minipage}[b]{0.065\textwidth}
			\includegraphics[width=1\textwidth]{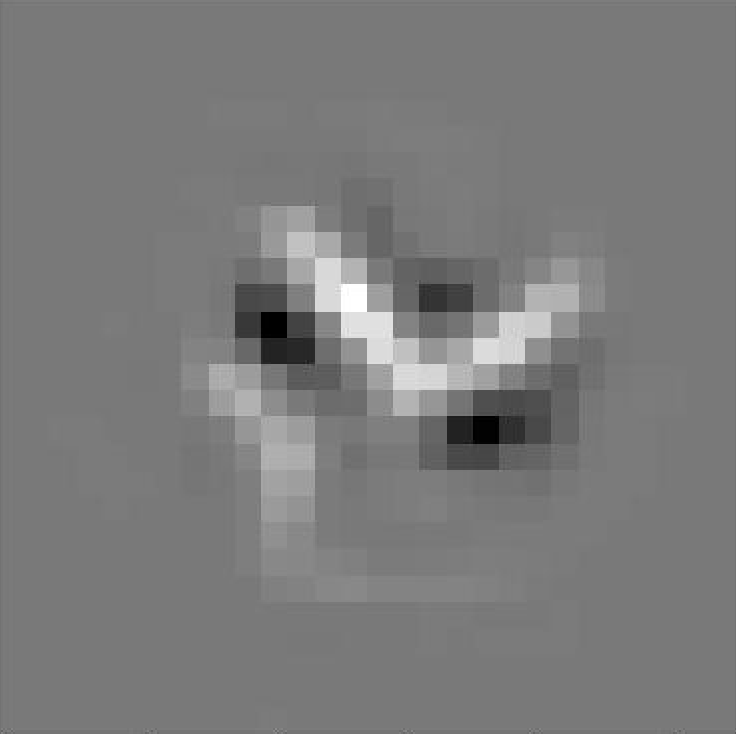} 
		\end{minipage}
		\label{fig:reverse_results_8}
	}
	\subfigure{
		\begin{minipage}[b]{0.065\textwidth}
			\includegraphics[width=1\textwidth]{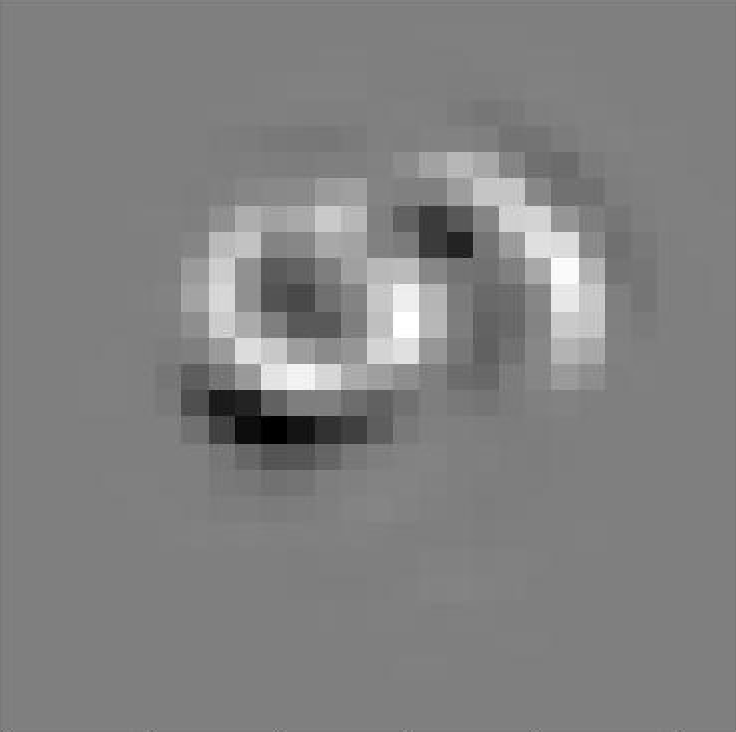} 
		\end{minipage}
		\label{fig:reverse_results_9}
	}
\caption{\ourmethod reverse engineering for 10 classes on MNIST. The true backdoor target is class 1. The first one is the clean image carrying the trigger. The rest are reversed triggers from class 0 to 9.}
\label{fig:reverse_results_10cl}
\end{figure}

\subsection{Discussion}
To explain our method, we analyze triggers reversed from every class under a simple case.
We use MNIST and a simple CNN architecture (see Appendix~\ref{sec:basicmodel} for details) with two convolutional layers and two fully connected layers.
We remove the constraint on the mask size to search for as powerful features as possible.
We replace $\mathcal{L}$ in Alg.~\ref{alg:updateuap} by: $\mathcal{L} = \mathcal{L}_{ce}(output, t) - SSIM(x, x').$
Under this setting, we train a backdoored model with BadNet.
Then, we conduct reverse engineering for all classes.
According to results in Fig.~\ref{fig:reverse_results_10cl}, the optimization with the loss $\mathcal{L}$ tends to learn unique class features for the clean class and the trigger features for the backdoor class.
This is expected as we only have a backdoor on the target class, i.e., class 1.
In this simplified situation, for clean classes without backdoors, only the unique class features allow the model to recognize that an input belongs to the class.
For the class injected with the backdoor, the model will recognize the input as the backdoor target based on the unique feature of the target and the feature of the backdoor trigger.

Reverse engineering requires a choice between the unique class feature and the feature of the backdoor trigger.
Regarding relatively simple features, the trigger feature is stronger than class features when training with poisoned data.
Reverse engineering can find a small perturbation with strong features enough to mislead the model according to learning objectives in the loss function.
However, in scenarios such as training with GTSRB or larger datasets, there might be strong features that can generate perturbations with a similar size to backdoor triggers.
This is why NC, Tabor, and \ourmethod provide more incorrect results in Tab.~\ref{tab:GTSRBdetection}, compared to results for MNIST and CIFAR-10.

\subsection{Details of the Basic Model}
\label{sec:basicmodel}
To reduce the impact of complex features and many model parameters, we use MNIST and a basic CNN architecture with two convolutional layers (followed by the ReLU activation function and a 2D average pooling layer) and two fully connected layers (denoted as the Basic model).
The input channel, output channel, and kernel size for the two convolutional layers are (1, 16, 5) and (16, 32, 5).
The input and output channels for the two fully connected layers are (512, 512) and (512, 10).
The model is trained using the same hyperparameters as Resnet-18 on MNIST in Section~\ref{sec:experimentmnist} with batch size=128, epochs=40, and poisoned rate=0.05.

\subsection{Datasets}
We use four popular datasets: MNIST~\cite{lecun1998mnist}, CIFAR-10~\cite{krizhevsky2009learning}, GTSRB~\cite{Stallkamp2012} and ImageNet~\cite{5206848}.

\begin{itemize}
    \item The MNIST contains 60,000 28$\times$28$\times$1 training images and 10,000 28$\times$28$\times$1 testing images in 10 classes.
    \item The CIFAR-10 contains 60,000 32$\times$32$\times$3 images in 10 classes, with 6,000 images per class, 50,000 for training, and 10,000 for testing.
    \item The GTSRB contains 51,840 traffic sign images in 43 categories. 39,210 for training and validation (80:20 split), 12,630 for testing (without labels). In experiments, the image size is fixed to 32$\times$32$\times$3.
    \item The ImageNet spans 1,000 classes and contains 1,281,167 training images, 50,000 validation images, and 100,000 test images. In experiments, the image size is fixed to 224$\times$224$\times$3.
\end{itemize}

\begin{figure*}[t]
\centering
\subfigure{
\rotatebox{90}{~~~~\footnotesize{NC}}
		\begin{minipage}[b]{0.08\textwidth}
			\includegraphics[width=1\textwidth]{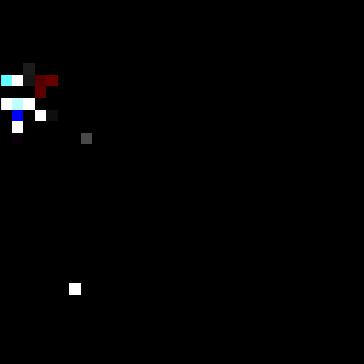} 
		\end{minipage}
	}
\subfigure{
		\begin{minipage}[b]{0.08\textwidth}
			\includegraphics[width=1\textwidth]{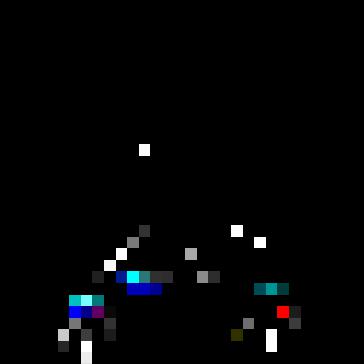} 
		\end{minipage}
	}
\subfigure{
		\begin{minipage}[b]{0.08\textwidth}
			\includegraphics[width=1\textwidth]{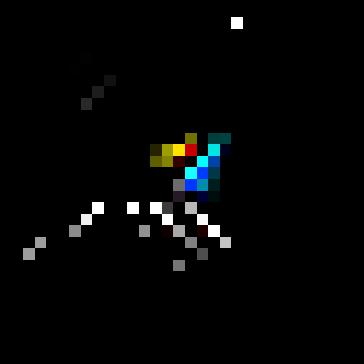} 
		\end{minipage}
	}
\subfigure{
		\begin{minipage}[b]{0.08\textwidth}
			\includegraphics[width=1\textwidth]{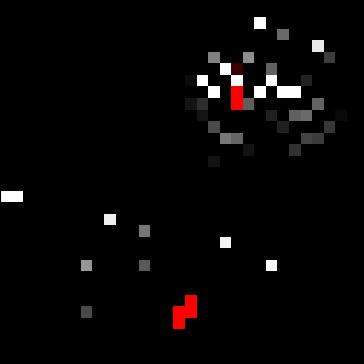} 
		\end{minipage}
	}
\subfigure{
		\begin{minipage}[b]{0.08\textwidth}
			\includegraphics[width=1\textwidth]{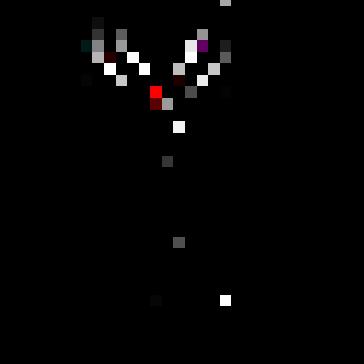} 
		\end{minipage}
	}
\subfigure{
		\begin{minipage}[b]{0.08\textwidth}
			\includegraphics[width=1\textwidth]{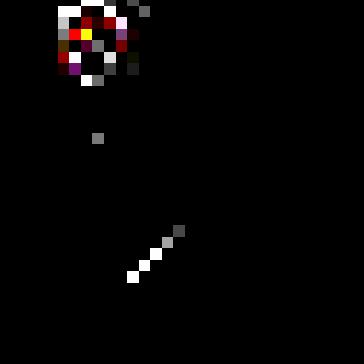} 
		\end{minipage}
	}
\subfigure{
		\begin{minipage}[b]{0.08\textwidth}
			\includegraphics[width=1\textwidth]{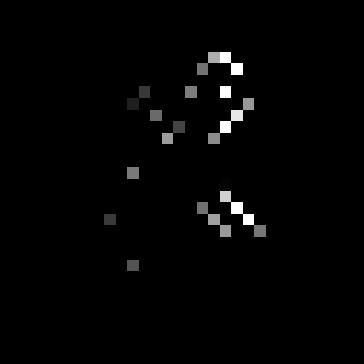} 
		\end{minipage}
	}
	\subfigure{
		\begin{minipage}[b]{0.08\textwidth}
			\includegraphics[width=1\textwidth]{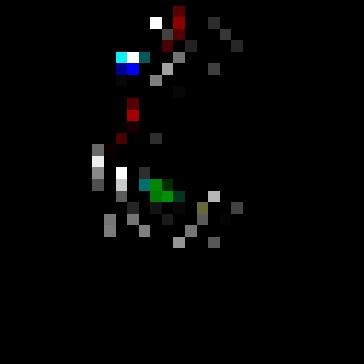} 
		\end{minipage}
	}
	\subfigure{
		\begin{minipage}[b]{0.08\textwidth}
			\includegraphics[width=1\textwidth]{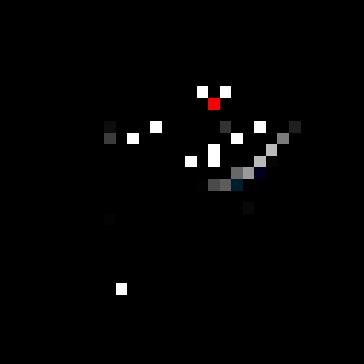} 
		\end{minipage}
	}
	\subfigure{
		\begin{minipage}[b]{0.08\textwidth}
			\includegraphics[width=1\textwidth]{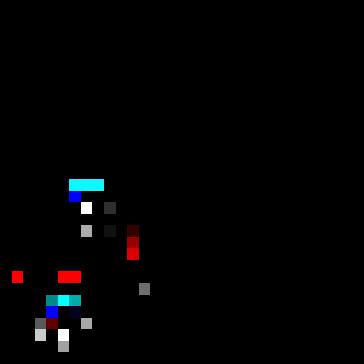} 
		\end{minipage}
	}\\
 \subfigure{
 \rotatebox{90}{\footnotesize{TABOR}}
		\begin{minipage}[b]{0.08\textwidth}
			\includegraphics[width=1\textwidth]{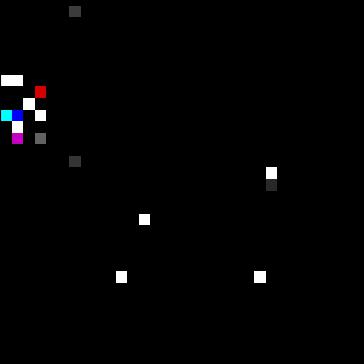} 
		\end{minipage}
	}
\subfigure{
		\begin{minipage}[b]{0.08\textwidth}
			\includegraphics[width=1\textwidth]{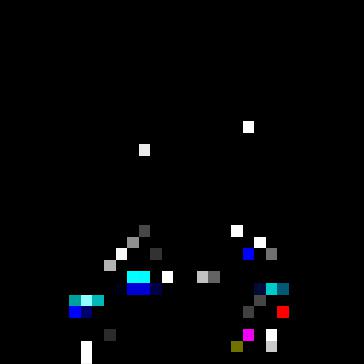} 
		\end{minipage}
	}
\subfigure{
		\begin{minipage}[b]{0.08\textwidth}
			\includegraphics[width=1\textwidth]{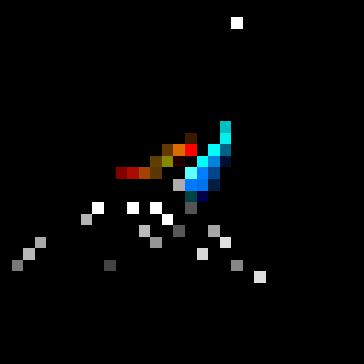} 
		\end{minipage}
	}
\subfigure{
		\begin{minipage}[b]{0.08\textwidth}
			\includegraphics[width=1\textwidth]{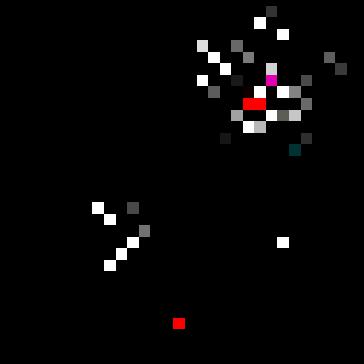} 
		\end{minipage}
	}
\subfigure{
		\begin{minipage}[b]{0.08\textwidth}
			\includegraphics[width=1\textwidth]{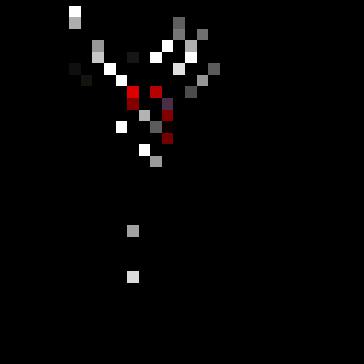} 
		\end{minipage}
	}
\subfigure{
		\begin{minipage}[b]{0.08\textwidth}
			\includegraphics[width=1\textwidth]{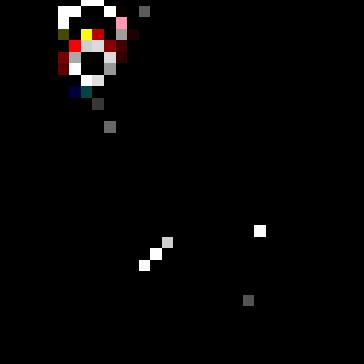} 
		\end{minipage}
	}
\subfigure{
		\begin{minipage}[b]{0.08\textwidth}
			\includegraphics[width=1\textwidth]{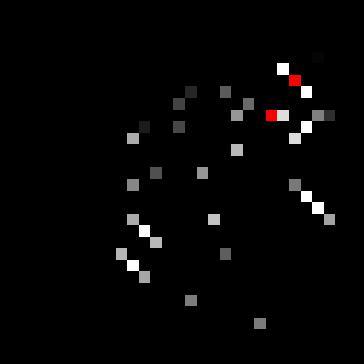} 
		\end{minipage}
	}
	\subfigure{
		\begin{minipage}[b]{0.08\textwidth}
			\includegraphics[width=1\textwidth]{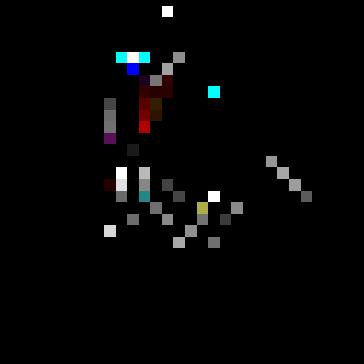} 
		\end{minipage}
	}
	\subfigure{
		\begin{minipage}[b]{0.08\textwidth}
			\includegraphics[width=1\textwidth]{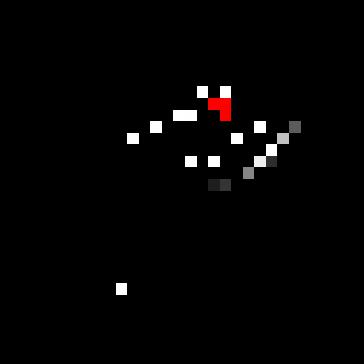} 
		\end{minipage}
	}
	\subfigure{
		\begin{minipage}[b]{0.08\textwidth}
			\includegraphics[width=1\textwidth]{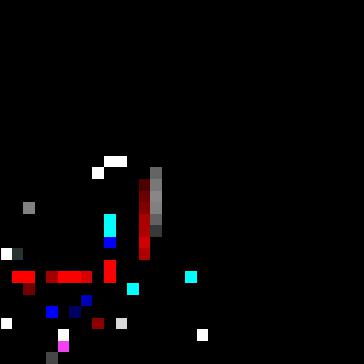} 
		\end{minipage}
	}\\
 \subfigure{
 \rotatebox{90}{~~~\ourmethod}
		\begin{minipage}[b]{0.08\textwidth}
			\includegraphics[width=1\textwidth]{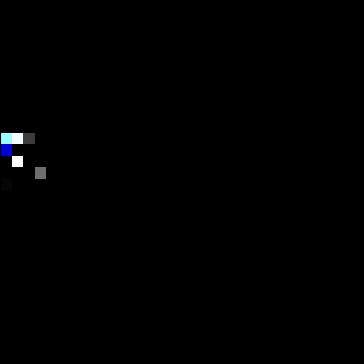} 
		\end{minipage}
	}
\subfigure{
		\begin{minipage}[b]{0.08\textwidth}
			\includegraphics[width=1\textwidth]{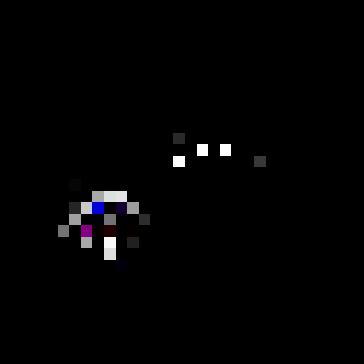} 
		\end{minipage}
	}
\subfigure{
		\begin{minipage}[b]{0.08\textwidth}
			\includegraphics[width=1\textwidth]{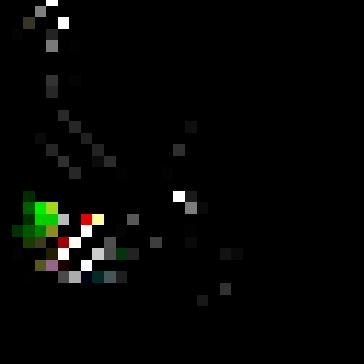} 
		\end{minipage}
	}
\subfigure{
		\begin{minipage}[b]{0.08\textwidth}
			\includegraphics[width=1\textwidth]{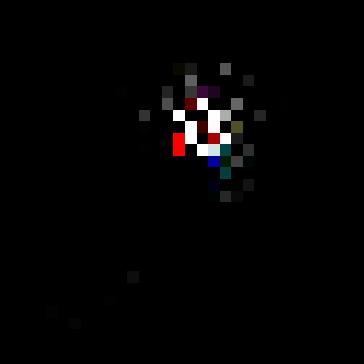} 
		\end{minipage}
	}
\subfigure{
		\begin{minipage}[b]{0.08\textwidth}
			\includegraphics[width=1\textwidth]{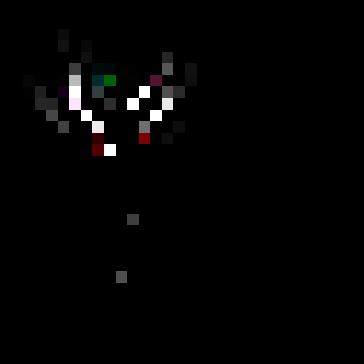} 
		\end{minipage}
	}
\subfigure{
		\begin{minipage}[b]{0.08\textwidth}
			\includegraphics[width=1\textwidth]{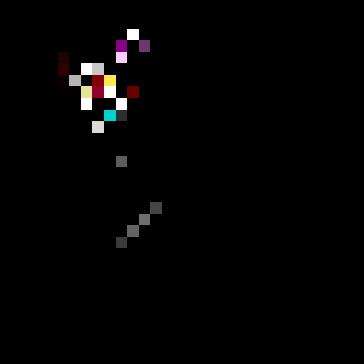} 
		\end{minipage}
	}
\subfigure{
		\begin{minipage}[b]{0.08\textwidth}
			\includegraphics[width=1\textwidth]{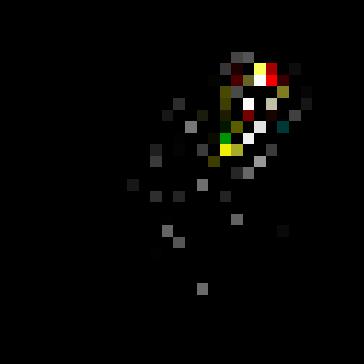} 
		\end{minipage}
	}
	\subfigure{
		\begin{minipage}[b]{0.08\textwidth}
			\includegraphics[width=1\textwidth]{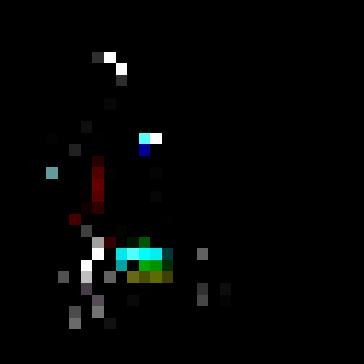} 
		\end{minipage}
	}
	\subfigure{
		\begin{minipage}[b]{0.08\textwidth}
			\includegraphics[width=1\textwidth]{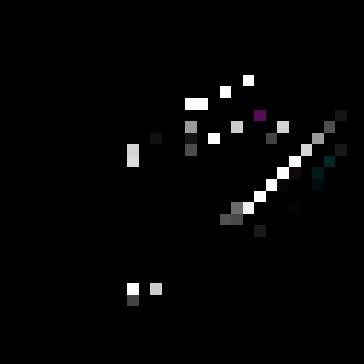} 
		\end{minipage}
	}
	\subfigure{
		\begin{minipage}[b]{0.08\textwidth}
			\includegraphics[width=1\textwidth]{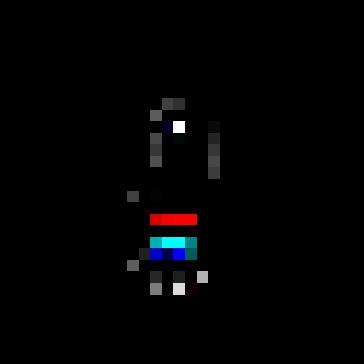} 
		\end{minipage}
	}
\caption{Reversed triggers from class 0 to 9.}
\label{fig:reverse_results_10cl_nc_tabor_usb}
\end{figure*}


\begin{table*}[tb]
\footnotesize
\centering
\setlength\tabcolsep{3pt}
\begin{tabular}{cccccccccccc}
\toprule
\multirow{2}{*}{Model} & \multirow{2}{*}{Method} & \multicolumn{10}{c}{GPU Time [m:s] in every class} \\
~& ~& 0 & 1 & 2 & 3 & 4 & 5 & 6 & 7 & 8 & 9\\
\midrule
\multirow{3}{*}{\tabincell{c}{Backdoored \\ (20 $\times$ 20 trigger)}} & NC & 23:16  & 24:18  & 24:32  & 23:39  & 24:48  & 23:35  & 23:15  & 23:34  & 23:41  & 24:10\\
~&TABOR& 33:54  & 37:24  & 34:19  & 35:51  & 33:59  & 36:45  & 34:23  & 36:47  & 35:4  & 36:23\\
~& \ourmethod & 4:26  & 4:26  & 4:27  & 4:30  & 4:26  & 4:26  & 4:26  & 4:26  & 4:26  & 4:26 \\
\midrule
\multirow{3}{*}{\tabincell{c}{Backdoored \\ (25 $\times$ 25 trigger)}} & NC & 23:35  & 24:38  & 25:11  & 24:3  & 24:58  & 24:19  & 24:29  & 23:54  & 24:29  & 23:6\\
~& TABOR& 48:23  & 47:24  & 48:48  & 48:41  & 48:38  & 47:41  & 48:27  & 49:10  & 48:48  & 47:45\\
~& \ourmethod& 4:44  & 4:42  & 4:44  & 4:44  & 4:49  & 4:48  & 4:48  & 4:54  & 4:50  & 4:45\\
\midrule
\multirow{3}{*}{\begin{minipage}[b]{0.06\columnwidth}
		\centering
		\raisebox{-.5\height}{\includegraphics[width=\linewidth]{././apple_white_tar0_alpha1.00_mark_30x30.png}}
	\end{minipage}}  & NC &19:1  & 18:22  & 20:47  & 18:5  & 18:48  & 21:27  & 18:54  & 18:30  & 20:26  & 17:57\\
~& TABOR& 48:52  & 47:16  & 49:0  & 48:39  & 48:54  & 47:40  & 48:34  & 48:55  & 48:55  & 47:50\\
~& \ourmethod& 4:27  & 4:26  & 4:27  & 4:30  & 4:26  & 4:26  & 4:26  & 4:26  & 4:26  & 4:25 \\
\bottomrule
\end{tabular}
\caption{Running time results of backdoor detection for Efficientnet-B0~\cite{pmlr-v97-tan19a}. Each result is the average of detection on 15 models.}
\label{tab:time_efficientnet}
\end{table*}

\end{document}